\documentclass[twoside,11pt]{article}
\usepackage{jair, theapa, rawfonts}
\usepackage{multirow}
\usepackage{longtable}
\usepackage{tikz}
\usepackage{tikz-qtree}
\usetikzlibrary{trees}
\usepackage{caption}
\usepackage{subcaption}
\usepackage{array}

\jairheading{1}{1993}{1-15}{6/91}{9/91}
\ShortHeadings{An Analysis of Deep Reinforcement Learning Agents for Text-based Games}
{ Chen, Dai, Poon \& Han}
\firstpageno{25}

\begin{document}

\title{An Analysis of Deep Reinforcement Learning Agents for Text-based Games}

\author{\name Chen Chen \email cche4088@uni.sydney.edu.au \\
       \name Yue Dai \email ydai7672@uni.sydney.edu.au \\
       \name Josiah Poon \email josiah.poon@sydney.edu.au \\
       \name Caren Han \email caren.han@sydney.edu.au,caren.han@uwa.edu.au \\
       \addr The University of Sydney\\
       Camperdown, NSW  2006 Australia
}


\maketitle

\begin{abstract}
Text-based games(TBG) are complex environments which allow users or computer agents to make textual interactions and achieve game goals.In TBG agent design and training process, balancing the efficiency and performance of the agent models is a major challenge. Finding TBG agent deep learning modules' performance in standardized environments, and testing their performance among different evaluation types is also important for TBG agent research. We constructed a standardized TBG agent with no hand-crafted rules, formally categorized TBG evaluation types, and analyzed selected methods in our environment.

\end{abstract}

\section{Introduction} \label{sec introduction}
Text-based games are designed for humans to play and finish quests using text replies. During one episode of a text-based game, when a player takes one action, the game environment provides some text descriptions about the current game state, typically about a place or some effects caused by the player's last action. Based on the text description(feedback), the player makes the next move, keeps collecting information, and tries to reach a winning state with the least steps possible.

\begin{figure*}[h]

        \centering
        \includegraphics[width=\linewidth]{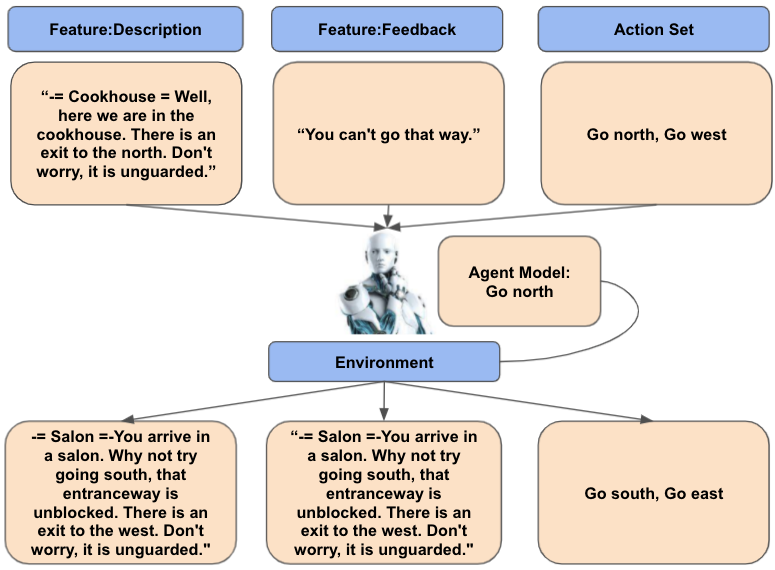}

    \caption{Text-based Game Agent Workflow}
    \label{figure:9}
\end{figure*}



Research on Text-based games mainly focuses on building goal-oriented smart agents. The methods in Text-based games could be applied to interactive language-related applications, such as the Internet of Things (IoT), smart assistants, online technical support/chatbots, and so on. The latest methods in this area include Reinforcement Learning and Natural Language Understanding modules. This combination enables smart agents to be trained through live language interactions and learn to understand the environment, improve themselves to perform actions and achieve goals.


The language models are typically trained offline in many traditional natural language tasks. However, in wide-ranging applications, like the smart agents in home pods, we expect the smart agent to adapt to different users in day-to-day interactions and figure out actions that need to take based on specific user language features and habits without hard-coded settings and improve user satisfaction. Text-based Games research is performed in a similar environment, where the agent explores the text environment, takes actions, collects results, and finally finds an optimal mapping of language input and actions to take without manually labelling the input data and training a model offline.

Some early papers in this field use hand-crafted rules as the core of the agents. As we expect Text-based game agents capable of online learning and improvement, this paper focuses only on Deep Reinforcement Learning agents. We also emphasize the deep learning module capability and efficiency analysis in this field, which is the critical component of agent online learning performance. The development path of this field is described in table \ref{table:1}. At the early years, the models were established on single-setting games, later with the development of Text-based Game experiment platforms. The agents are trained on multiple games with different settings. More methods that have been successfully applied to the Natural Language Processing field are introduced into Text-based game agents, including graphs and external knowledge.

The most regularly used platform in this field is Textworld. Due to the complexity of Text-based game agents, even trained on the same platform, hand-crafted rules/language features and other component differences make it hard to compare the effectiveness of deep learning module performance alone. To offer a clearer view of method effectiveness among benchmarked works, We collected some deep learning modules and methods from related research works, benchmarked them under the same environments, and experimented with the same hyperparameters in different game scenarios. 

Some new approaches to the Text-based game task have shown outstanding results, like graph-aided techniques. Rather than learning games in one set, or using as many features as possible, new challenges in this field includes using less hand-crafted features and training the agent to solve unseen games.

The motivation for this survey is to examine literature in this field after 2015, analyze it in depth and benchmark some methods in a standardized environment. We summarize approaches introduced so far, present our benchmark results and analysis, and point out some future directions. In this paper, we consider three questions:
\begin{itemize}
    \item Q1: How do different agent models balance efficiency and performance?
    \item Q2: How do some modules perform on an agent without hand-crafting rules and features in standardized environments?
    \item Q3: Can models perform consistently in different evaluation types?
\end{itemize}




\subsection{Related analysis and surveys}

Most of the Text-based Games agent field methods are evaluated in separate research works. Though each research work selects benchmarks to present the significance of the findings, the evaluations are conducted with different games and settings. They might incorporate different levels of hand-crafted rules, with additional modules on top of the Deep Neural Network structure. Several surveys in this field collected methods and results within the research field. Inspired by these works, we decided to focus our analysis on deep learning modules used in TBG agents, review the related methods more in-depth, test selected models in a standardized evaluation setting and present our results and analysis.

\subsection{Structure of the analysis}
This analysis begins with an overview of the literature and related methodology. We introduce the methodology development trend in this field, review and classify the methods, and explain how they are applied. We review the evaluation platforms and metrics used in related research works and then introduce our ablation test on selected methods within a standardized environment. Finally, we summarize our analysis of the methods and findings and address the issues in this field for future direction.

\section{Overview of the literature} \label{sec overview}


Text-based game agents are designed to explore the game environment, extract useful information from text features provided, and learn optimal policy to give correct text actions. Efforts have been made to improve text encoding components, processing text features, and efficiently producing text actions. Issues are learning global policy, accurately constructing game state representations, and reducing state space.


Text-based game is generally framed as a reinforcement learning problem. In early years, many hand-crafted methods are introduced to try to solve games by analyzing text patterns and crafting templates. Deep Reinforcement learning is later introduced into this field. Test environments like Textworld are introduced to enable learning from a batch of games, and testing agent generalization ability on unseen games. Neural networks such as MLP, RNN are applied into agent design as text encoder and being developed over ensuing years.











\begin{table*}[t!]
\centering
\caption{ Historical overview of text-based game papers}

\begin{tabular}{  m{0.8cm}  m{6.5cm} m{6.5cm}  } 

\textbf{Year} & \textbf{Network Architecture/Training} & \textbf{Reinforcement Learning Methods}  \\

2015 &  MLP, Word Embeddings & State Encoding, Deep Q-Learning  \\

2018 & State Recurrence, Multiple games, Zero-shot Evaluation & Exploration Reward, Replay Memory  \\ 

2019 & Graph Attention, pre-trained knowledge graph & Graph Representation, Actor-Critic 








\end{tabular}
\label{table:2}
\end{table*}



Table\ref{table:2} summarizes major methodologies introduced into this field, ranked by timeline. The two earliest papers\cite{narasimhan2015language,he2015deep} (2015, table \ref{table:2}) introducing deep reinforcement learning into text-based games are inspired by NLP methods. Popularly methods like word embeddings and MLP are introduced for agents to process and encode text descriptions. A word embedding sequence represents the game state, and vanilla DQN is used as RL algorithm. With the two earliest works, more variations of neural network architecture, such as RNN and Transformer are introduced. The actor-critic algorithm was introduced as an alternative Reinforcement Learning framework to DQN. Various graph methods are also introduced as a new type of state representation. More pre-trained modules are included in text-based game agents for enhancement.

Text-based game platforms are also under continuous development. The earliest works only focus on one particular game. Later platforms allow the evaluation to be done on different games. On top of different games, Textworld\cite{cote2018textworld}(2018,table\ref{table:2}) offers a large batch generation of similar games, which allow agents to be trained on a large training set to allow testing on unseen games.

The latest development of text-based game agent research is more focused on a graph representation of game state(2019,table\ref{table:2}), and several graph generation methods are introduced. Recently, researchers also focus on enhancing agent language understanding capability by including pre-trained language processing components into agents' neural network architecture.

\section{Reinforcement Learning Methods}

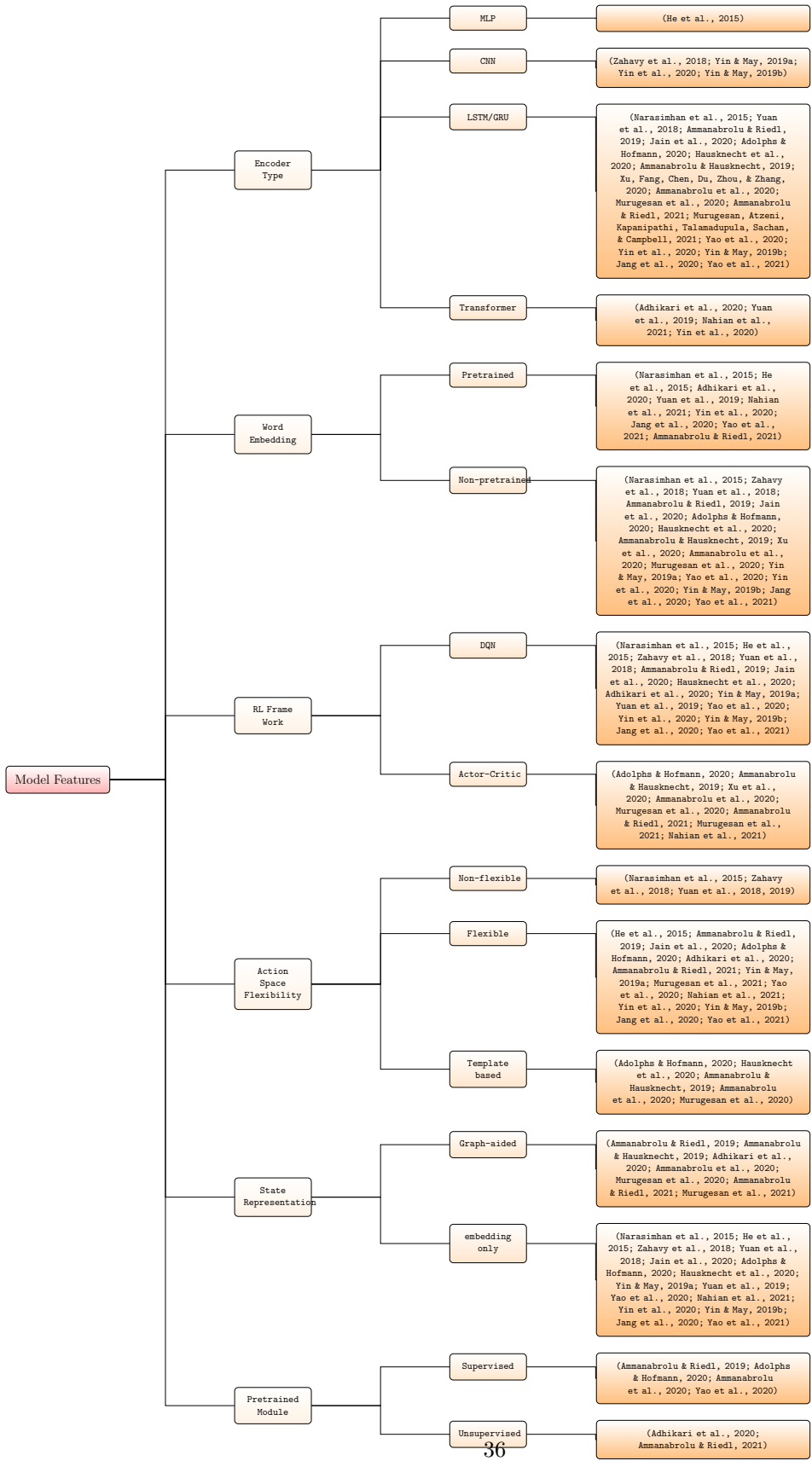
\begin{figure}
\begin{tikzpicture}[grow'=right, level distance=3.3in, sibling distance=.25in,scale=0.4,
treenode/.style = {shape=rectangle, rounded corners,
                     draw, align=center,
                     top color=white, bottom color=blue!20,inner sep=10pt},
  edge from parent/.style= 
            {thick, draw, edge from parent fork right},
  root/.style     = {treenode, font=\Large, bottom color=red!30},
  l1/.style      = {treenode, font=\ttfamily\normalsize, minimum width=1.1in,text width=0.9in, bottom color=orange!10},
  l2/.style      = {treenode, font=\ttfamily\normalsize, minimum width=1.0in,text width=0.9in, bottom color=orange!20},
  l3/.style      = {treenode, font=\ttfamily\normalsize, minimum width=3.0in,text width=3.0in, bottom color=orange!50}
  ]

\Tree 
    [. \node[root] {Model Features}; 
        [. \node [l1] {Encoder Type};
            [. \node[l2] {MLP};
              [. \node[l3] {\cite{he2015deep}};
              ]
            ]
            [. \node[l2] {CNN}; 
              [. \node[l3] {\cite{zahavy2018learn,yin2019comprehensible,yin2020learning,yin2019learn}};
              ]
            ]
            [. \node[l2] {LSTM/GRU}; 
              [. \node[l3] {\cite{narasimhan2015language,yuan2018counting,ammanabrolu2019playing,jain2020algorithmic,adolphs2020ledeepchef,hausknecht2020interactive,ammanabrolu2019graph,xu2020deep2,ammanabrolu2020avoid,murugesan2020enhancing,ammanabrolu2021learning,murugesan2021efficient,yao2020keep,yin2020learning,yin2019learn,jang2020monte,yao2021reading}};
              ]
            ]
            [. \node[l2] {Transformer}; 
              [. \node[l3] {\cite{adhikari2020learning,yuan2019interactive,nahian2021training,yin2020learning}};
              ]
            ]
        ]
        [. \node [l1] {Word Embedding};
            [. \node[l2] {Pretrained};  
              [. \node[l3] {\cite{narasimhan2015language,he2015deep,adhikari2020learning,yuan2019interactive,nahian2021training,yin2020learning,jang2020monte,yao2021reading,ammanabrolu2021learning}};
              ]
            ]
            [. \node[l2] {Non-pretrained}; 
              [. \node[l3] {\cite{narasimhan2015language,zahavy2018learn,yuan2018counting,ammanabrolu2019playing,jain2020algorithmic,adolphs2020ledeepchef,hausknecht2020interactive,ammanabrolu2019graph,xu2020deep2,ammanabrolu2020avoid,murugesan2020enhancing,yin2019comprehensible,yao2020keep,yin2020learning,yin2019learn,jang2020monte,yao2021reading}};
              ]
            ]
        ]
        [. \node [l1] {RL Frame Work};
            [. \node[l2] {DQN}; 
              [. \node[l3] {\cite{narasimhan2015language,he2015deep,zahavy2018learn,yuan2018counting,ammanabrolu2019playing,jain2020algorithmic,hausknecht2020interactive,adhikari2020learning,yin2019comprehensible,yuan2019interactive,yao2020keep,yin2020learning,yin2019learn,jang2020monte,yao2021reading}};
              ]
            ]
            [. \node[l2] {Actor-Critic}; 
              [. \node[l3] {\cite{adolphs2020ledeepchef,ammanabrolu2019graph,xu2020deep2,ammanabrolu2020avoid,murugesan2020enhancing,ammanabrolu2021learning,murugesan2021efficient,nahian2021training}};
              ]
            ]
        ]
        [. \node [l1] {Action Space Flexibility};
            [. \node[l2] {Non-flexible}; 
              [. \node[l3] {\cite{narasimhan2015language,zahavy2018learn,yuan2018counting,yuan2019interactive}};
              ]
            ]
            [. \node[l2] {Flexible}; 
              [. \node[l3] {\cite{he2015deep,ammanabrolu2019playing,jain2020algorithmic,adolphs2020ledeepchef,adhikari2020learning,ammanabrolu2021learning,yin2019comprehensible,murugesan2021efficient,yao2020keep,nahian2021training,yin2020learning,yin2019learn,jang2020monte,yao2021reading}};
              ]
            ]
            [. \node[l2] {Template based}; 
              [. \node[l3] {\cite{adolphs2020ledeepchef,hausknecht2020interactive,ammanabrolu2019graph,ammanabrolu2020avoid,murugesan2020enhancing}};
              ]
            ]
        ]
        [. \node [l1] {State Representation};
            [. \node[l2] {Graph-aided}; 
              [. \node[l3] {\cite{ammanabrolu2019playing,ammanabrolu2019graph,adhikari2020learning,ammanabrolu2020avoid,murugesan2020enhancing,ammanabrolu2021learning,murugesan2021efficient}};
              ]
            ]
            [. \node[l2] {embedding only}; 
              [. \node[l3] {\cite{narasimhan2015language,he2015deep,zahavy2018learn,yuan2018counting,jain2020algorithmic,adolphs2020ledeepchef,hausknecht2020interactive,yin2019comprehensible,yuan2019interactive,yao2020keep,nahian2021training,yin2020learning,yin2019learn,jang2020monte,yao2021reading}};
              ]
            ]
        ]
        [. \node [l1] {Pretrained Module};
            [. \node[l2] {Supervised};
              [. \node[l3] {\cite{ammanabrolu2019playing,adolphs2020ledeepchef,ammanabrolu2020avoid,yao2020keep}};
              ]
            ]
            [. \node[l2] {Unsupervised}; 
              [. \node[l3] {{\cite{adhikari2020learning,ammanabrolu2021learning}}};
              ]
            ]
        ]
    ]
\end{tikzpicture}
\caption{Ontology for TBG agent methods}
\label{ontology}

\end{figure}

As text-based game agent training requires interaction with the game environment, most related research works are built under the Reinforcement Learning(RL) framework. Various RL methods have been applied, including game states and action interpretation, RL algorithms and reward functions. In this section, we introduce basic concepts of major methodologies and classify related works in an ontology (Figure \ref{ontology}) for reader's reference.

\subsection{State Representation}
Under the Reinforcement Learning framework, a text-based game environment can be modelled as POMDP, and the agents do not have direct access to the game states. One of the challenges for solving text-based games, even for a human, is interpreting visible game features and choosing the correct data formation to represent the game's state as accurately as possible.

\subsubsection{Embedding only}
Research works like \cite{narasimhan2015language,he2015deep} only use word embedding sequences generated from game state text description as game representations. Pre-trained, and non-pretrained methods are adopted, which is discussed in the word embedding section. Some researchers also concatenate features like game file names as game state representations.
\subsubsection{Graph aided}
Graph has been widely used as a feature to solve problems in many fields, such as bioinformatics and computer vision. For text-based games like Textworld games, the game states can be accurately represented by the graph. The accurate graph representation is usually not accessible to game players. Using pretrained neural networks or hand-crafted rules, some researchers try turning game state text description and agent experience into knowledge graphs. Neural network architectures like GCN can encode the generated knowledge graph.  

\subsection{Action Space}
Action space flexibility describes a text-based game agent's ability to adapt to complex action text and action sets. Some agents can only cope with a small fixed action set, and others can deal with action text with varying lengths, large vocabulary, and complex combinations.
\subsubsection{Non-flexible}
Text-based game agents' deep learning models could be designed with a fixed output fit a small action set. For example, \cite{yuan2018counting} fixed their network output shapes to 2 and 5, to deal with a fixed 2*5 (10 different action choices in total) game action set. This design has different sets of neural network layer weights for different action choices, which could be beneficial to predict value more accurately for each action. Yet due to the network output size only fits a fixed group of actions, this design is hard to generalize to games with longer action text sentences and varying action choices.
\subsubsection{Flexible}
To deal with action text sentences with arbitrary length and larger vocabulary, agent neural network models could be designed to have only one output node to predict state-action values. This kind of model is designed to predict values for all available actions and choose one to act based on the policy. These models have a stronger ability to generalize to different games. Yet, it doesn't have other groups of weight trained for each action, which makes the training process harder than non-flexible type models.
\subsubsection{Template-based}
Research works such as \cite{zahavy2018learn,jain2020algorithmic} have text action templates based on human gameplay experience. They only use the deep learning model to predict appropriate words to fill in the text action templates. This model type shares the same generalization problem with the non-flexible type, and the whole template needs to be redesigned to generalize to different games. Also, these models will need human gameplay experience to determine the most suitable text templates.

\subsection{Reinforcement Learning Algorithm}
Reinforcement Learning frameworks such as policy gradient, Q-learning and Actor-Critic are widely adopted in game solving and bioinformatics fields. \cite{narasimhan2015language} firstly adopted Deep Q-Learning to solve text-based games, and some following works adopted the same framework. Actor-critic is a widely used algorithm in recent years and has been successfully applied to complex problems like Go game.

\subsubsection{Deep Q-Learning}
Deep Q-learning(DQN) is a less complex yet powerful Reinforcement Learning algorithm. DQN is widely used in solving games like Atari. Q-learning uses the below function to update Q value and solve the game by approximating the Q value for each state.
\begin{equation}
Q_(s_{t},a_{t}) = \gamma\max_{a_{t+1}}Q(s_{t+1},a_{t+1})+r_{t}
\end{equation}

where $s_{t},a_{t},r_{t}$ demotes state,action and reward at time step $t$. 

For the text-based games field, \cite{narasimhan2015language} firstly adopted DQN to approximate the Q value for state-action and state-object pairs and select optimal pairs as the input for each text game state description. \cite{yuan2018counting} also used DQN, and their result showed that, with some additional reward design, DQN could support the text-based games to solve complicated games and generalize the performance on unseen games.

\subsubsection{Actor-Critic}
Actor-critic is considered a more advanced algorithm than DQN, and its capability has been proven to solve very complex games such as the GO game. Researchers have used Advantage Actor Critic(A2C) to train text-based game agents. A2C use below formula to calculate value \(v(s_{t})\) ,return \(R(s_{t},a_{t}) \) and advantage \(Adv(s_{t},a_{t}) \)for game states and actions. 
\begin{equation}
R(s_{t},a_{t}) = \gamma^{l}v(s_{t+l})+\sum_{k=1}^{l} \gamma^{k-1}r_{t+k}\
\end{equation}
\begin{equation}
Adv(s_{t},a_{t}) = R(s_{t},a_{t})-v(s_{t})
\end{equation}
 
 A2C is considered a reliable agent training method as it reduces variances using advantage. Yet most TBG research work doesn't compare performance between applying Q-learning and A2C, as it requires neural network architecture change.

\subsubsection{Reward Functions}

The goal of a Text-based game agent is to collect game scores from the environment as much as possible. A game score can be directly adapted as a reward to the agent. Researchers also designed reward functions to encourage certain agent behaviours, such as choosing actions likely to lead the agent to unexplored game states. For example, the exploratory reward function designed by \cite{yuan2018counting} encourages the agent to explore unseen states.


\section{Deep Learning Architectures and modules} \label{sec joint}

\subsection{Encoder Type}
As text-based game agents need to process text information given by game environment, the architecture of the text encoder type have a large influence on agent performance. The agent model also needs to be lite, as the agents are trained by sampling from game environments, and the text encoder needs to be efficient enough to support agent exploration.

\subsubsection{MLP}
Multilayer Perceptron(MLP) is a feed-forward neural network consisting of an input layer, one or more hidden layers and an output layer. MLP is a basic form of neural network and is widely used in domains like classification, regression and reinforcement learning. \cite{he2015deep} uses MLP to encode text inputs in TBGs. For each state/action text pair, DRRN\cite{he2015deep} feeds the state embedding vectors and action embedding vectors into two MLPs separately, then approximate the Q value based on the inner product of vectors from the last hidden layers of both MLPs.
\subsubsection{CNN}
Because of its outstanding performance in deep learning, Convolutional Neural Network(CNN) has become one of the most popular neural networks. Although CNN is more popular in computer vision, its performance in other domains like time series prediction and signal identification is also noticeable. The basic architecture of CNN includes an input layer, convolutional layers, pooling layers, fully-connected layers and an output layer. Various CNN models like VGGNet\cite{simonyan2015deep}, GoogLeNet\cite{szegedy2015going} and ResNet\cite{he2015deep} differ in the number of layers, number and size of kernels, etc.. For TBG tasks, \cite{zahavy2018learn,yin2019comprehensible,yin2020learning,yin2019learn} leverage CNN to encode game description text input. All the models are modified based on \cite{kim-2014-convolutional}. \cite{zahavy2018learn,yin2019comprehensible} use the similar CNN structure in \cite{kim-2014-convolutional}. \cite{yin2020learning,yin2019learn} only use a lite version of CNN, which consists of an input layer, Convolutional layer and max-pooling layer. The output of the max-pooling layer is the encoding of text input. Besides, \cite{yin2019comprehensible,yin2019learn,yin2020learning} add position embedding into CNN input to assist model learning.
\subsubsection{LSTM/GRU}
A recurrent Neural Network(RNN) is widely used in processing sequential data. It is commonly used in tasks, like machine translation, sentiment analysis, video analysis, speech recognition, etc. Long Short-Term Memory(LSTM) and Gated Recurrent Unit(GRU) are two typical forms of RNN. The input of both LSTM and GRU is word embedding, and each cell generates an output and a hidden state passed to the next cell. Research works such as\cite{narasimhan2015language,yuan2018counting} tokenize game state description text, map one-hot embeddings to higher dimensional word embedding sequences, and use LSTM to encode the embedding sequence as state encoding. Other works such as \cite{adolphs2020ledeepchef,hausknecht2020interactive} utilize GRU. Both output of LSTM/GRU and the hidden state of the last cell can be used as the game state encoding. 
\subsubsection{Transformer}
Since Transformer\cite{vaswani2017attention} was proposed, it has quickly become the hot field in research because of its great success. Although the original transformer is designed for NLP tasks, variants of transformers have entered other domains like computer vision and audio processing. The transformer in \cite{vaswani2017attention} has an encoder and a decoder. Both encoder and decoder take the word embedding sequence generated from game state text, and then add positional embedding onto the word embedding sequence. This sequence is used as the transformer encoder input. For TBG field, in\cite{adhikari2020learning}, \cite{yuan2019interactive} and \cite{yin2020learning} use transformer encoder to encode game state description text, and use the transformer encoder output as state or action encoding.
\subsubsection{BERT}
Bidirectional Encoder Representations from Transformers(BERT)\cite{kenton2019bert} is a pre-trained transformer-based model. It achieves many state-of-art performances in NLP tasks. In the TBG research field, research works like\cite{nahian2021training,yin2020learning} use BERT for state description text encoding. In these works, part of the weights of the encoder is frozen, and only a small part of the encoder layers will be fine-tuned during training, which makes the training more efficient. Compared to the Transformer encoder, the pre-trained BERT model normally uses it's own tokenizer and produces word segment embeddings. It has a much more number of parameters to capture information from a huge corpus. Among all the encoder types, BERT is the heaviest model.
\subsection{Word Embedding}

Word embeddings are used to represent tokens in given texts. Word embeddings can be pre-trained from the selected corpus and fixed for later use. Researchers sometimes unfreeze pretrained embedding weights to let downstream tasks update them. Most text-based game agents use embedding sequences to represent the game state or action text.

\subsubsection{Pretrained}
\cite{adhikari2020learning}, \cite{yuan2019interactive},\cite{nahian2021training} and \cite{yin2020learning} leverage pre-trained word embedding models and keep the embeddings fixed during training the agent. Among them, \cite{adhikari2020learning} and \cite{yuan2019interactive} use fastText\cite{mikolov2018advances}, \cite{yin2020learning} use Glove\cite{pennington2014glove} and BERT, \cite{nahian2021training} use BERT. \cite{jang2020monte} use spaCy\cite{Honnibal_spaCy_Industrial-strength_Natural_2020} which is a NLP library that provides static pre-trained word embeddings. In addition, \cite{he2015deep} uses bag-of-words(BOW), \cite{narasimhan2015language} compares BOW and bag-of-bigrams(BI) representation with trained LSTM embedding. \cite{yao2021reading} uses hash values as word embedding and shows the agent can achieve a high score even without language semantics.
\subsubsection{Non-pretrained}
Most of the agents discussed in this paper either fine-tune the pre-trained word embeddings or train the embeddings from scratch. Research works like \cite{narasimhan2015language}, \cite{yuan2018counting} initialize the word embeddings as either zeros or random vectors. Some research works like \cite{yao2021reading} and  \cite{murugesan2020enhancing} initialize the embedding using pre-trained Glove model, \cite{zahavy2018learn} uses word2vec\cite{mikolov2013efficient} model instead. Based on the experiment from \cite{yin2020learning}, although fully fine-tuning BERT outperforms freezing it during training, the former sacrifices the latter's advantage in training speed. \cite{yao2020keep} conducts similar experiments, and they compare fine-tuning pre-trained GPT-2\cite{radford2019language}---a model that only consists of decoder part of transformer---and learning a random initialized GPT-2, the result shows pre-trained GPT-2 has better performance.

\subsection{Pretrained Modules}
Pretrained modules are widely used in text-based game agent designs. With pretrained modules, text-based game agents are initialized with knowledge before exploring game environments. Using pretrained modules can improve agent performance, yet potentially could damage agent efficiency and generalization ability towards different games.
\subsubsection{Supervised}
\cite{ammanabrolu2019playing} trains a paired question encoder and answer encoder using DrQA\cite{chen2017reading} method. The data are generated from an agent that is accessible to the shortest solution of the game. The weights of the encoders and the embedding learned are used to initialize the agent. Ammanabrolu et al.\cite{ammanabrolu2019playing} only tested this method on TextWorld games with a 'home' theme in this paper. Later, they extended the question-answering pairs to a set of Jericho games and proposed Jericho-QA\cite{ammanabrolu2020avoid}. In 
this paper, Ammanabrolu et al.\cite{ammanabrolu2020avoid} pre-trained ALBERT\cite{lan2019albert} with Jericho-QA. The information extracted by ALBERT from text observation is used to update the knowledge graph during training. \cite{adolphs2020ledeepchef} pre-trained a task-specific module called Recipe Manager. It is used to predict what's left to complete the tasks. As mentioned, it can only be used in the cooking games generated by TextWorld. \cite{yao2020keep} pre-trains CALM with ClubFloyd--a dataset containing state-action pairs of TBGs--- to generate a relatively small size of possible actions at each turn of the game. 
\subsubsection{Unsupervised}
\cite{adhikari2020learning} proposed two methods to pre-train knowledge graph updater, which are Observation Generation (OG) and Contrastive Observation Classification (COC). OG trains the graph updater with a decoder to reconstruct text observation. COC instead trains the model to differentiate between true observation and a fake one.

\begin{figure}
    \centering
    \includegraphics[width=\linewidth]{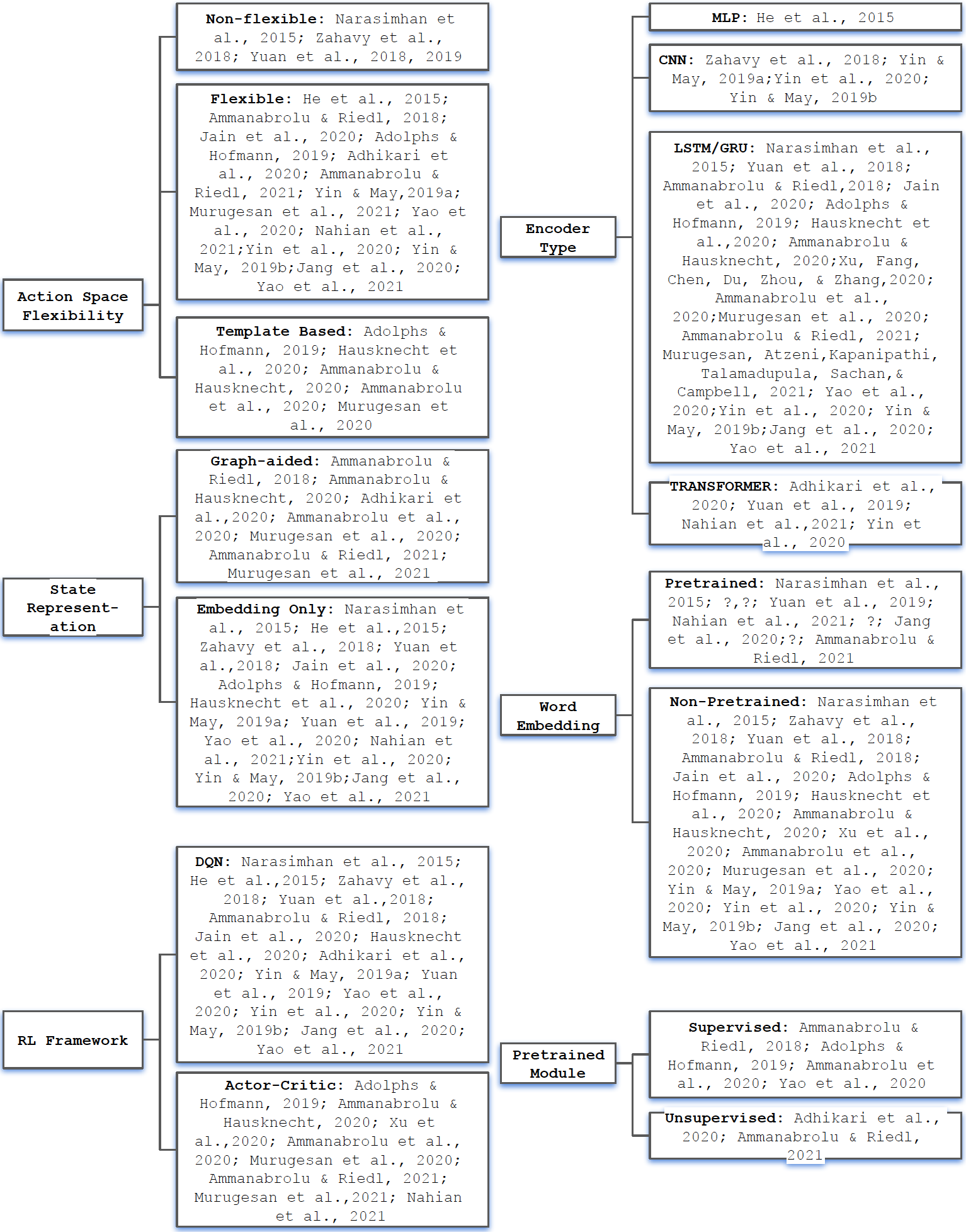}
    \caption{ontology}
    \label{figure:ontology}
\end{figure}



\section{Benchmark Environments}

\subsection{Text-based game platforms}
\subsubsection{Single games}
At an early stage in the Text-based Games agent research field, researchers like \cite{narasimhan2015language}\cite{he2015deep} use single game environments like Evennia and Ifarchive. The games typically offer limited difficulty settings and a few fixed game maps. For this kind of environment, the agent only needs to learn how to solve a single game, which Reinforcement Learning tabular methods can handle, as state and action combinations are very limited.
\subsubsection{Text-based game platforms}
In 2018, Cote et al.\cite{cote2018textworld} built the first platform to support text-based game research. The environment supports building a set of games with more detailed language descriptions, which enables agents to be trained to understand the language and solve a class of games. Hausknecht\cite{hausknecht19} developed the Jericho platform, included additional features like games state trees, and connected it to the Textworld platform.

Urbanek et al.\cite{urbanek2019light} developed a Light environment and combined the game with dialogue. The games in this platform do not have complicated rooms and tasks like Cook in Textworld, but it supports training agents that perform tasks and chat at the same time.

\subsection{Text-based game agent evaluation settings}

Text-based game agent learning and evaluation settings can be classified as follows:\\
\textbf{1.Single game} The agent learns how to play one game and achieve more game scores possible. Some single games with very large maps and multiple subtasks can be hard to deal with, like Zork.\\
\textbf{2.Multiple games} The agent learns to solve a set of games with similar difficulty and settings. Each game in the set can have different objects and room configurations, and the text description for rooms can also be different. Still, the general task and difficulty are very similar in the game set. An agent is evaluated with average performance among all the games in the set.\\
\textbf{3. Generalisation} The agent learns from a set of games for training and is evaluated by the performance on the training set and on some unseen games to test if the learnt policy generalises to unseen language descriptions and settings.\\

\begin{figure}

\begin{tikzpicture}[grow'=right,level distance=3.1in,sibling distance=1in,scale=0.7]
\tikzset{
  treenode/.style = {shape=rectangle, rounded corners,
                     draw, align=center,
                     top color=white, bottom color=blue!20, level distance = 3em,},
  edge from parent/.style= 
            {thick, draw, edge from parent fork right},
  root/.style     = {treenode, font=\Large, bottom color=red!30},
  l1/.style      = {treenode, font=\ttfamily\normalsize, minimum width=0.9in,text width=1.2in, bottom color=orange!10},
  l2/.style      = {treenode, font=\ttfamily\normalsize, minimum width=0.9in,text width=3in, bottom color=orange!20}
  }
  
\Tree 
    [. \node[root] {Evaluation Types}; 
            [. \node[l1] {single game}; 
              [. \node[l2] {\cite{narasimhan2015language,he2015deep,zahavy2018learn,yuan2018counting,jain2020algorithmic,hausknecht2020interactive,ammanabrolu2019graph,ammanabrolu2020avoid,murugesan2020enhancing,ammanabrolu2021learning,yin2019comprehensible,nahian2021training,jang2020monte,yao2021reading}};
              ]
            ]
            [. \node[l1] {multiple games}; 
              [. \node[l2] {{\cite{yuan2018counting,ammanabrolu2019playing,jain2020algorithmic,adolphs2020ledeepchef,adhikari2020learning,murugesan2020enhancing,ammanabrolu2021learning,yuan2019interactive,murugesan2021efficient,yao2020keep,yin2019learn}}};
              ]
            ]
            [. \node[l1] {generalization}; 
              [. \node[l2] {{\cite{yuan2018counting,adolphs2020ledeepchef,adhikari2020learning,yuan2019interactive,yin2020learning,yin2019learn}}};
              ]
            ]
        ]
    ]
\end{tikzpicture}
\caption{Evaluation Types of related works}
\label{Evaluation types}
\end{figure}
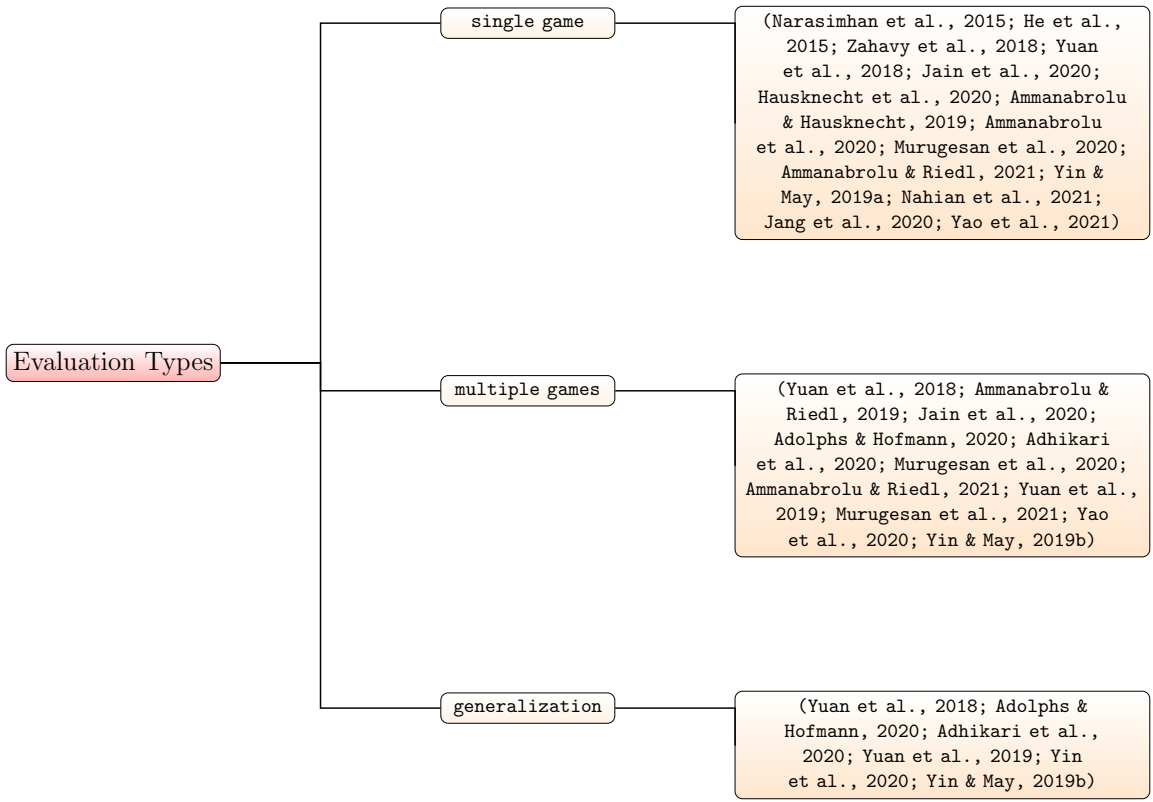

Most of the methods in this field are only evaluated in limited settings, and we classify the evaluation methods used in related works in ontology figure \ref{Evaluation types}. 

\begin{figure}
    \centering
    \includegraphics[width=0.5\linewidth]{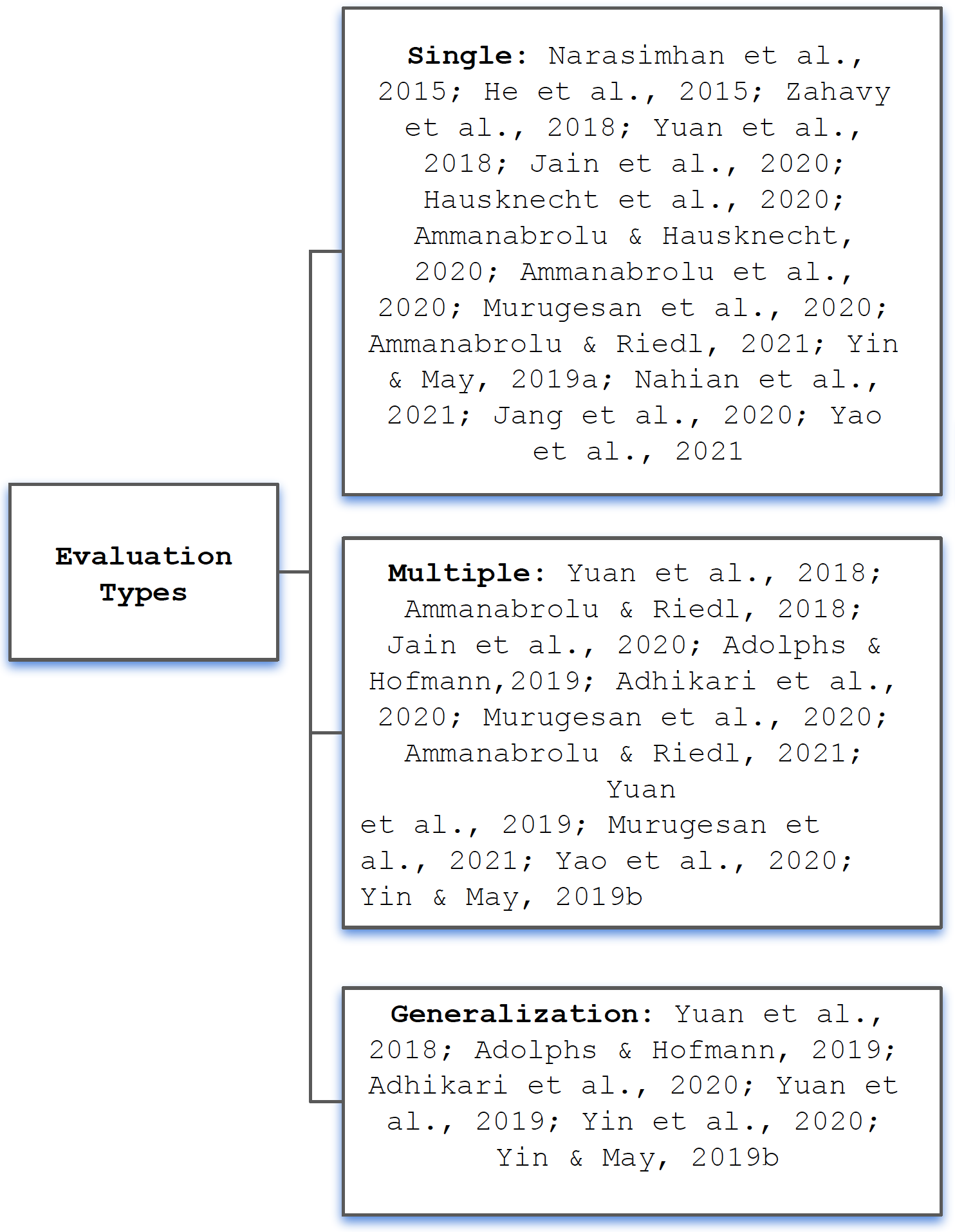}
    \caption{ontology}
    \label{figure:ontology}
\end{figure}

\subsection{Evaluation metrics}
Two metrics are generally adopted in this text based game agent evaluation: \\
\textbf{(1)} The average game completion rate:$\sum(s_{o})/(s_{m}*b)$, where$\hspace{1mm}s_{o}=\hspace{1mm}$obtained score by the agent in a game,$\hspace{1mm}s_{m}=\hspace{1mm}$maximum game score, $\hspace{1mm}b=\hspace{1mm}$evaluation set size;\\
\textbf{(2)} the average steps for the agent to finish games: $\sum(s_{p})/b$, where$\hspace{1mm}s_{p}=\hspace{1mm}$step taken for the agent to finish a game. 









\section{Ablation Test and Analysis} \label{sec metrics}

To evaluate and analyse methods used in the Text-based Games research field, particularly for the methods' performance in solving multiple games and unseen games, we created a standardised environment and agents and performed ablation testing for selected methods.

\subsection{Ablation Test Experiment settings}
We evaluate selected models in several Textworld\cite{cote2018textworld} environments. Textworld is the most commonly used platform in this field. As it supports large game set generation, it enables us to train models on large game sets, perform generalisation tests and see how the testing models perform on unseen games. We selected three games: Coin Collector, Treasure Hunter and Cooking recipe. The chosen games have different levels of difficulty. Coin collector is the easiest, and it has very few rooms($\sim$20), a simple task, and a fixed action set. The Cooking recipe game's task is more challenging, and the agent needs to collect multiple items and perform cooking actions to finish the game, the action set of different states could be different. Treasure hunter's difficulty is between Coin Collector and Cooking recipe.

We report the models' performance in two different settings: multiple games and generalisation: \\
\textbf{(1)} For multiple games setting, we develop and test different games and train/test our model on the set of games; \\
\textbf{(2)} To test model performance on unseen games, we evaluate the models in a generalisation setting. In this setting, we train the models by using 100 generated games and test the models on 20 games that the models have never seen during training. \\

The training and testing set is formed by different games that are randomly chosen from a larger game set. For the generalisation test, we report both training and testing results. To better visualise the data changing trend in the figures, the values are filtered with Savgol filter\cite{savitzky1964smoothing} to reduce noise.

  




\begin{table}[h!]
  
  \label{tab:commands}
  \begin{tabular}{p{0.3\linewidth}  p{0.2\linewidth} p{0.2\linewidth}}

    \textbf{Feature} & \textbf{State Number} & \textbf{Action Voc Size}\\

    Coin Collector Lv5 & 308 & 10\\ 

    Treasure Hunter Lv5 & 426  & 24\\

    Cooking Recipe lv1 & 374 & 32\\

  \end{tabular}
  \caption{Game Environment statistics}
  \label{table:1}
\end{table}

\subsection{Ablation Test Methods}
\subsubsection{Encoder types}

For encoder types, we benchmark five different kinds of text encoder architectures popularly used by related works. The five encoders are chosen to cover the main classes of Neural Network architectures and, at the same time, cover different parameter number levels so we can analyze the balance of performance and efficiency. We selected GRU and LSTM as our test encoders for Recurrent Neural Networks, which adopt a timestep recurrence methodology. These two models are the most commonly used RNNs. GRU has a less complex structure than LSTM, as RL training requires better efficiency for agent exploration. We want to compare the performance of GRU and LSTM and see if GRU works faster with a less complex structure. The rest three encoders(CNN,BERT,Transformer) are non-recurrent type encoders. CNN adopts a convolution methodology, while Transformer and BERT use an attention mechanism. BERT is the most popular model in traditional NLP tasks, yet it has many more parameters than the others. 

\begin{figure*}[h]
    \centering
    \begin{subfigure}[b]{0.48\textwidth}
        \centering
        \includegraphics[width=\linewidth]{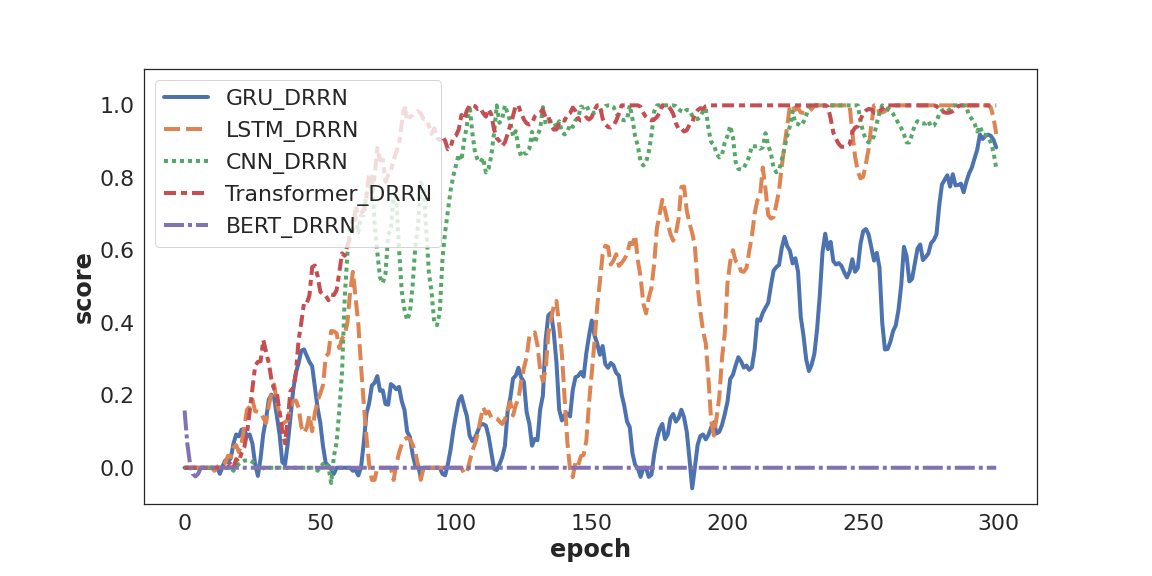}
        \caption{Average training score}
    \end{subfigure}
    \begin{subfigure}[b]{0.48\textwidth}
        \centering
        \includegraphics[width=\linewidth]{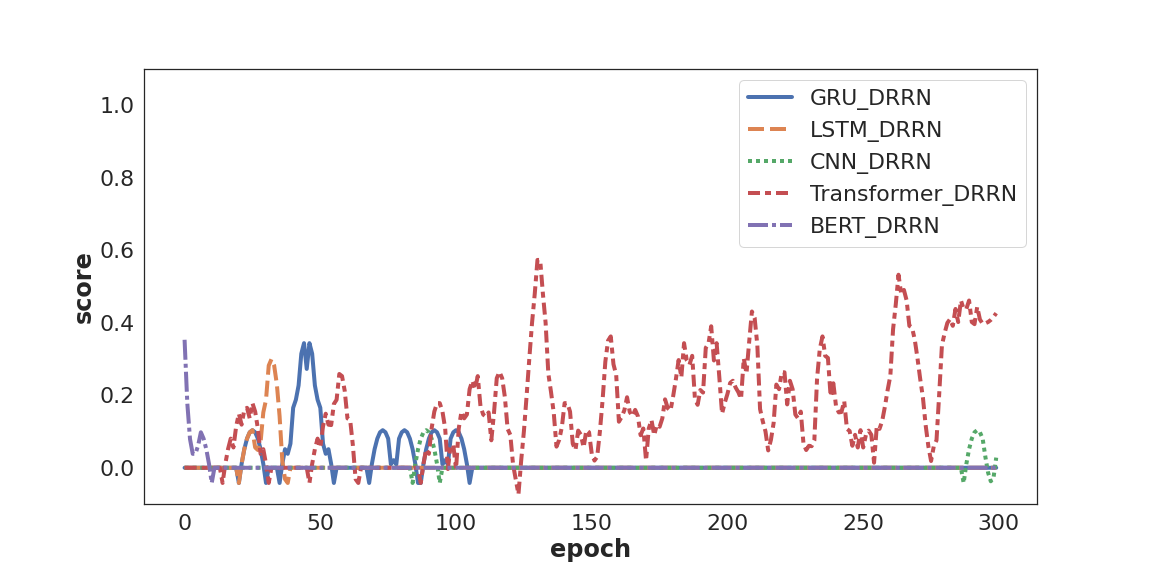}
        \caption{Average evaluation score}
    \end{subfigure}
    \caption{Average scores for different encoder types in Coin Collector game}
    \label{figure:1}
\end{figure*}

\begin{figure*}
    \centering
    \begin{subfigure}[b]{0.48\textwidth}
        \centering
        \includegraphics[width=\linewidth]{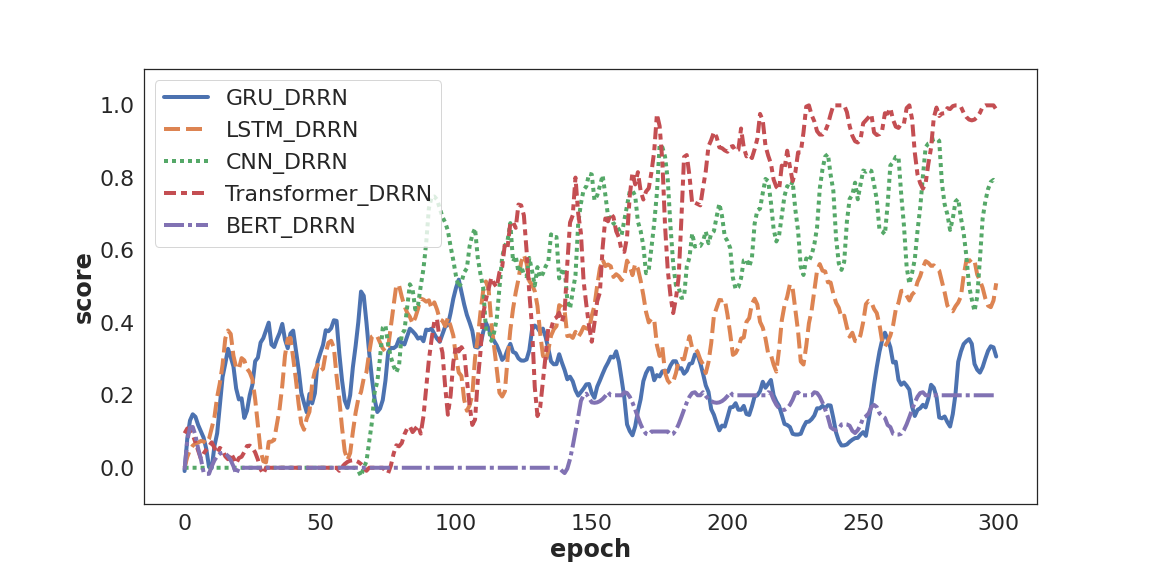}
        \caption{Average training score}
    \end{subfigure}
    \begin{subfigure}[b]{0.48\textwidth}
        \centering
        \includegraphics[width=\linewidth]{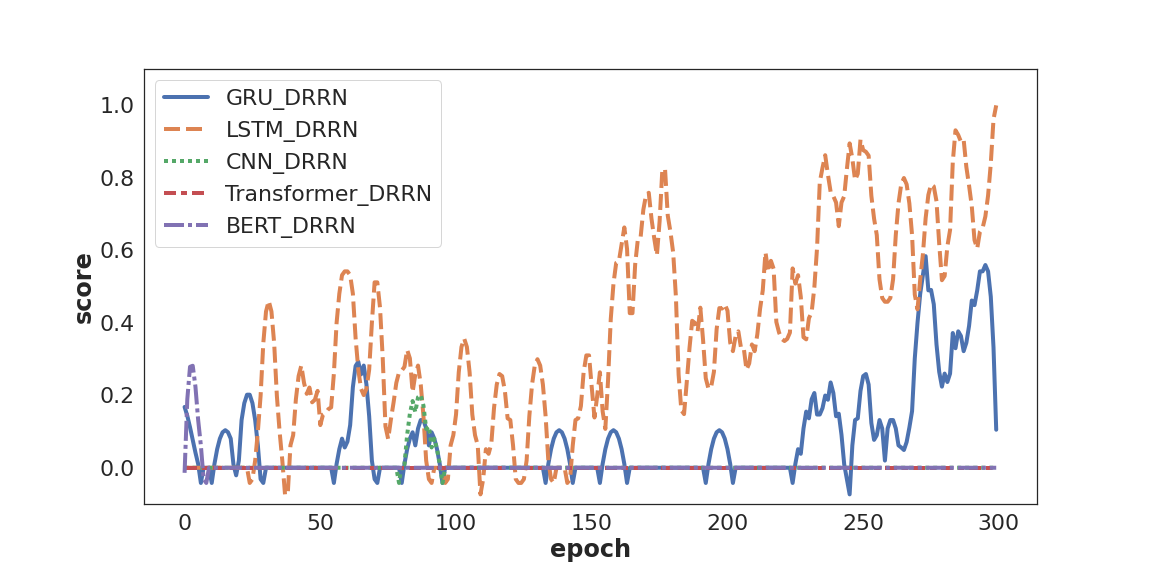}
        \caption{Average evaluation score}
    \end{subfigure}
    \caption{Average scores for different encoder types in Treasure Hunter game}
    \label{figure:2}
\end{figure*}

\begin{figure*}[h]
    \centering
    \begin{subfigure}[b]{0.48\textwidth}
        \centering
        \includegraphics[width=\linewidth]{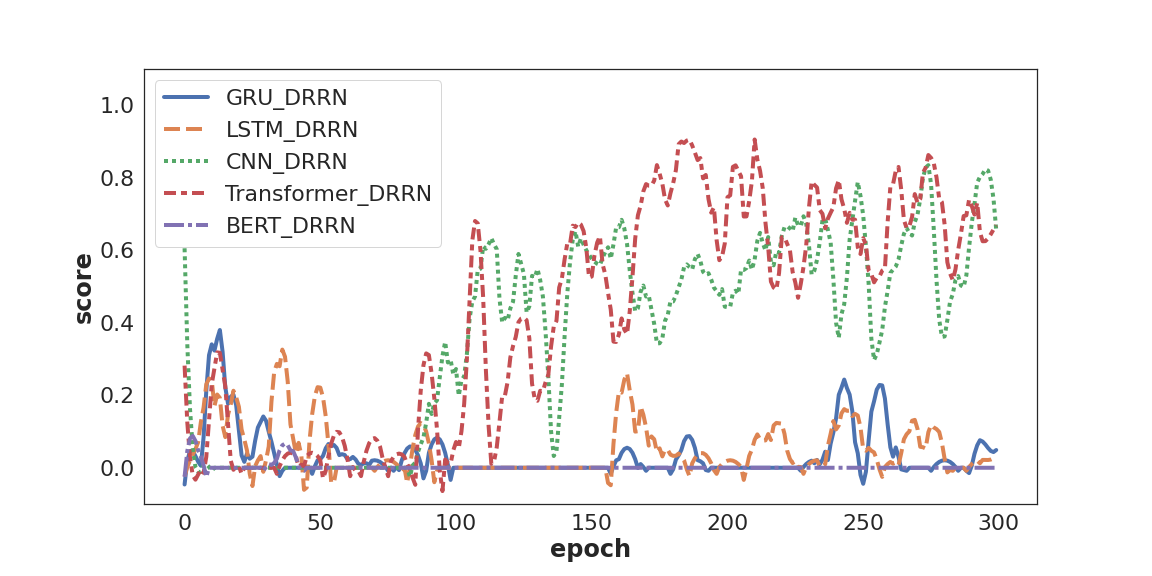}
        \caption{Average training score}
    \end{subfigure}
    \begin{subfigure}[b]{0.48\textwidth}
        \centering
        \includegraphics[width=\linewidth]{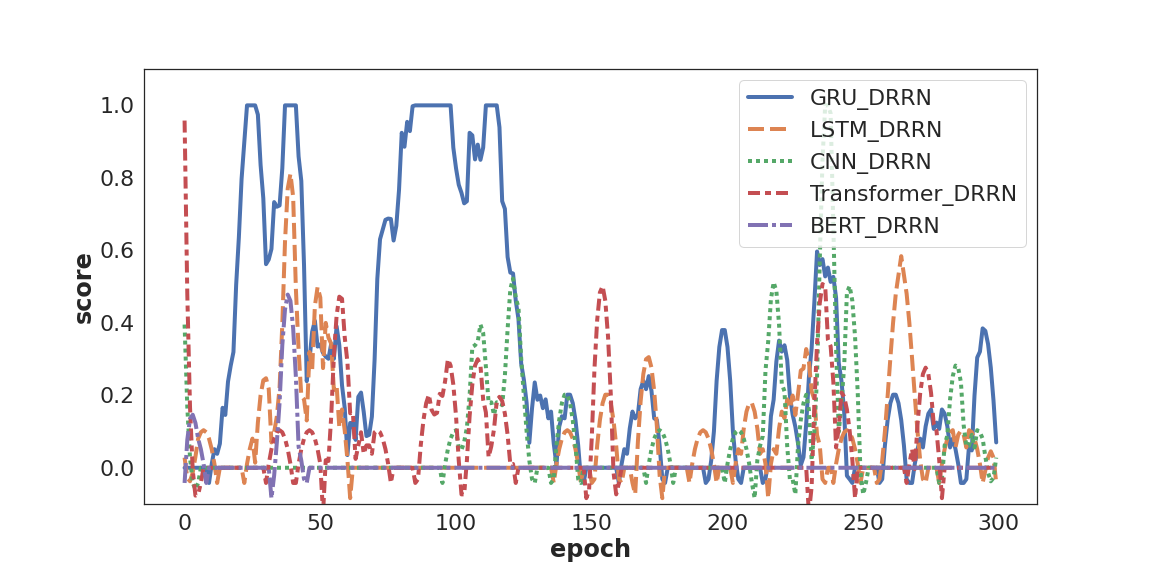}
        \caption{Average evaluation score}
    \end{subfigure}
    \caption{Average scores for different encoder types in Cook game}
    \label{figure:3}
\end{figure*}

From ablation test results on a training set with 100 different games reported in fig \ref{figure:1}-\ref{figure:3}(a), the CNN encoder learns faster than other models. The transformer encoder has comparable performance to CNN encoder, especially in the Cook game. Both CNN and Transformer encoder converge for the Coin-collection game and Treasure hunter game. CNN encoder learns slightly faster. The two models are powerful enough to learn useful information in the Cook game, which is very challenging to solve, without any additional modules.For RNN models, we tested GRU and LSTM encoder. Two encoders have similar performance in all settings but can only converge to optimal policy in Coin-Collector. The two recurrent encoders failed to converge for the other two more complex games. For the Treasure hunters, they can learn some useful information and solve some games in the set, but in the Cook game, they can't stably learn from the environment. We also tested the BERT encoder, which is an SOA language learning architecture. Surprisingly, without pretrained weights, BERT can barely learn anything from the environment, which indicates that model behaviour can be very different when learning in interactive environments like Text-based games, compared to learning from the corpus for traditional NLP tasks.

We also performed a generalisation test by evaluating trained agents on 20 unseen games. The different encoders' performance on other games is inconsistent Fig.\ref{figure:1}(b) shows that the Transformer agent has outstanding performance on overlooked Coin Collector games, Fig.\ref{figure:2}(b) show that RNN encoders perform better on unseen Treasure hunter games. In fig.\ref{figure:3}(b), except BERT encoder, all other encoder types have similar performance. Compared to solving a fixed set of games, agent performance on unseen games is much lower (generally lower than 0.4), and no obvious pattern can be observed from the evaluation scores. Solving unseen games remains a significant challenge in TBG agent research field.\\

As an RL agent must explore and learn from the environment efficiently, we also report Encoder parameters statistics. BERT has the largest parameter size, yet it's unable to learn effectively from all the environment settings. The model can converge if we use a pretrained BERT and freeze most of its layers. In this case, the model needs to be pre-trained with a prepared corpus. We deem this a sort of word embedding and will discuss this in the word embedding section. BERT is unsuitable for TBG tasks if additional training corpus or pre-trained parameters are unavailable.

GRU/LSTM/Transformer encoders have similar parameters count. The Transformer outperforms GRU and LSTM significantly. CNN has a slightly better performance compared to Transformer, yet it has 7.7 times more parameters. The agent needs to sample and explore the environment for complex TBG tasks to converge to the optimal policy. Using the same training time, agents with a Transformer encoder can explore much more states than agents with a CNN encoder. Parameter counts are summarised in the below list.

\begin{itemize}
  \item \textbf{GRU} 1,255,169
  \item \textbf{LSTM} 1,288,193
  \item \textbf{CNN} 13,469,377
  \item \textbf{Transformer} 1,741,041
  \item \textbf{BERT} 109,942,017
\end{itemize}

\subsubsection{Word Embeddings}

\begin{figure*}[h]
    \centering
    \begin{subfigure}[b]{0.48\textwidth}
        \centering
        \includegraphics[width=\linewidth]{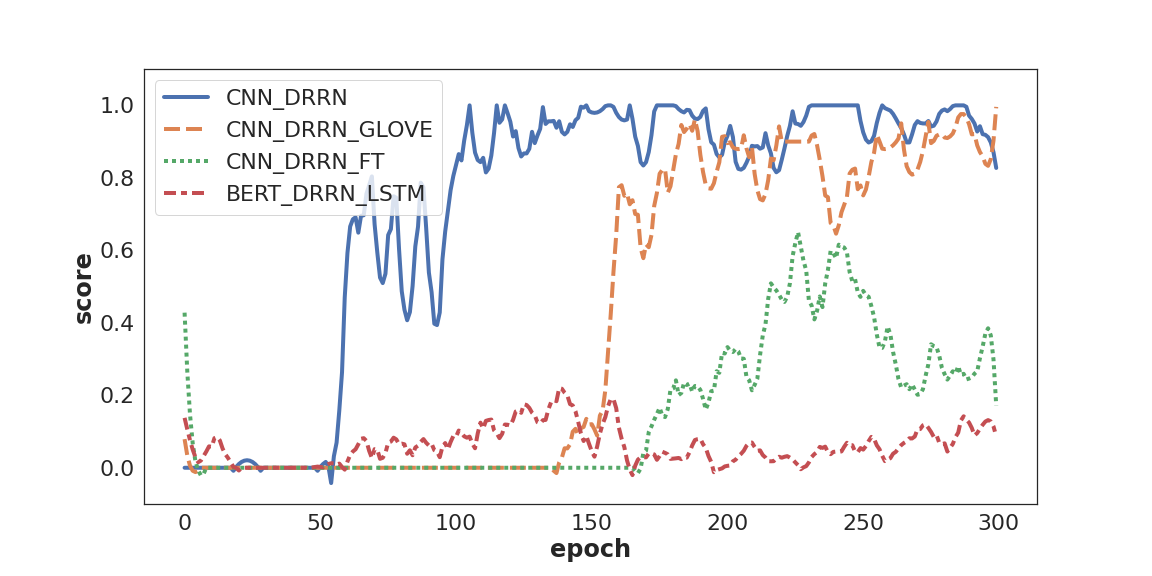}
        \caption{Average training score}
    \end{subfigure}
    \begin{subfigure}[b]{0.48\textwidth}
        \centering
        \includegraphics[width=\linewidth]{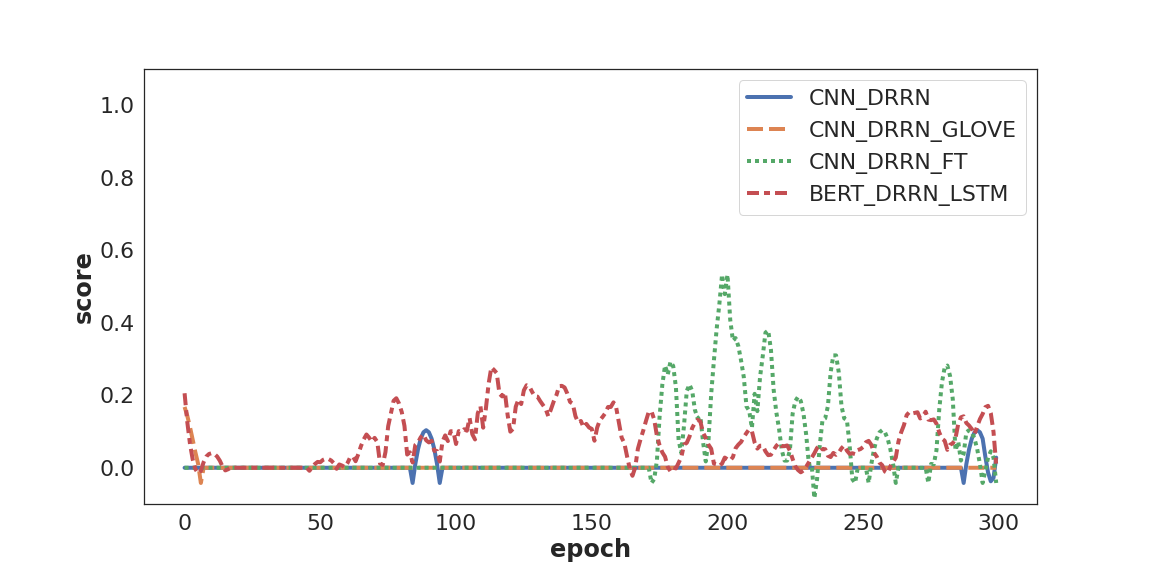}
        \caption{Average evaluation score}
    \end{subfigure}
    \caption{Average scores for GRU encoder with different embedding types}
    \label{figure:5}
\end{figure*}

\begin{figure*}[h]
    \centering
    \begin{subfigure}[b]{0.48\textwidth}
        \centering
        \includegraphics[width=\linewidth]{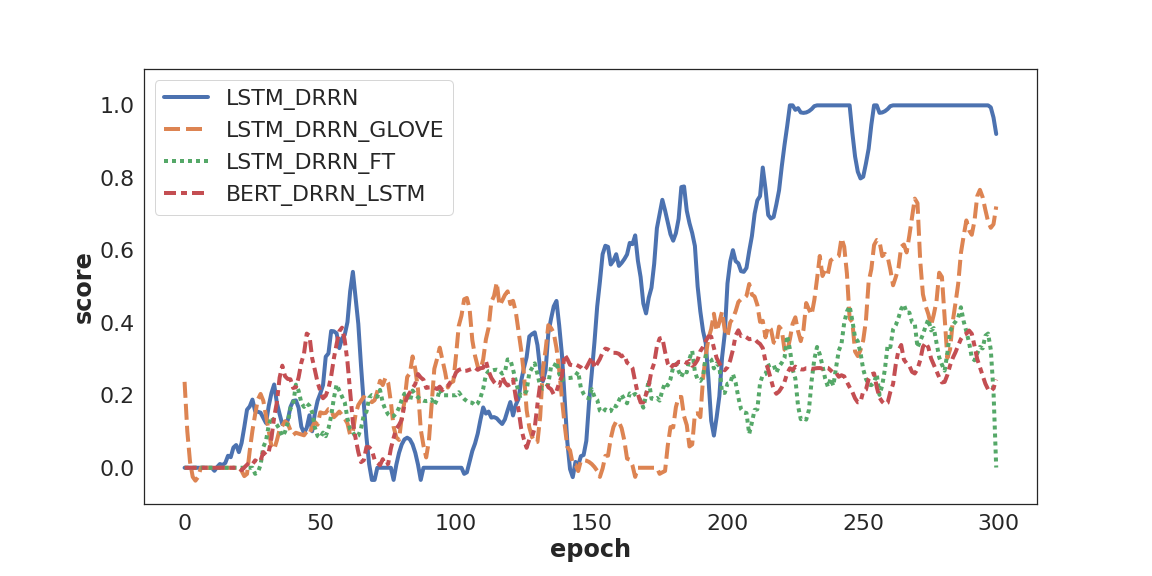}
        \caption{Average training score}
    \end{subfigure}
    \begin{subfigure}[b]{0.48\textwidth}
        \centering
        \includegraphics[width=\linewidth]{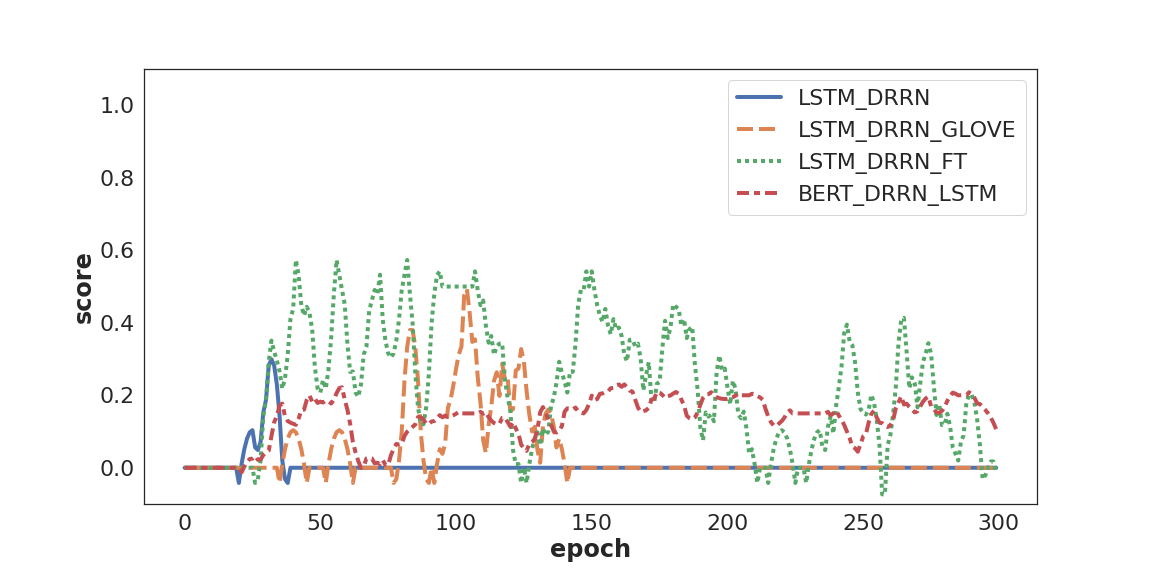}
        \caption{Average evaluation score}
    \end{subfigure}
    \caption{Average scores for LSTM encoder with different embedding types}
    \label{figure:6}
\end{figure*}

\begin{figure*}[h]
    \centering
    \begin{subfigure}[b]{0.48\textwidth}
        \centering
        \includegraphics[width=\linewidth]{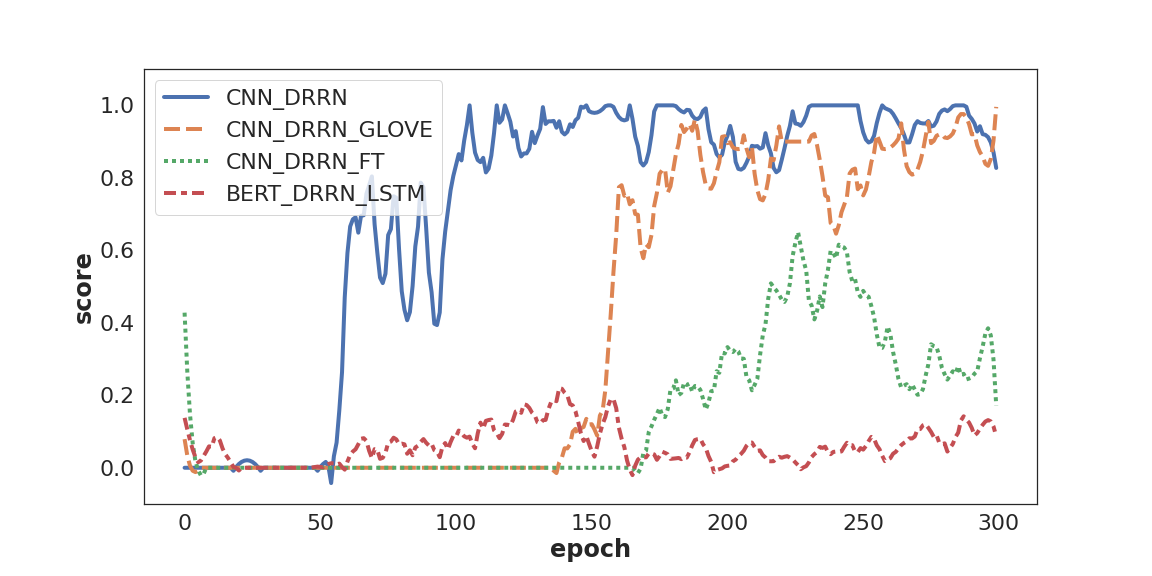}
        \caption{Average training score}
    \end{subfigure}
    \begin{subfigure}[b]{0.48\textwidth}
        \centering
        \includegraphics[width=\linewidth]{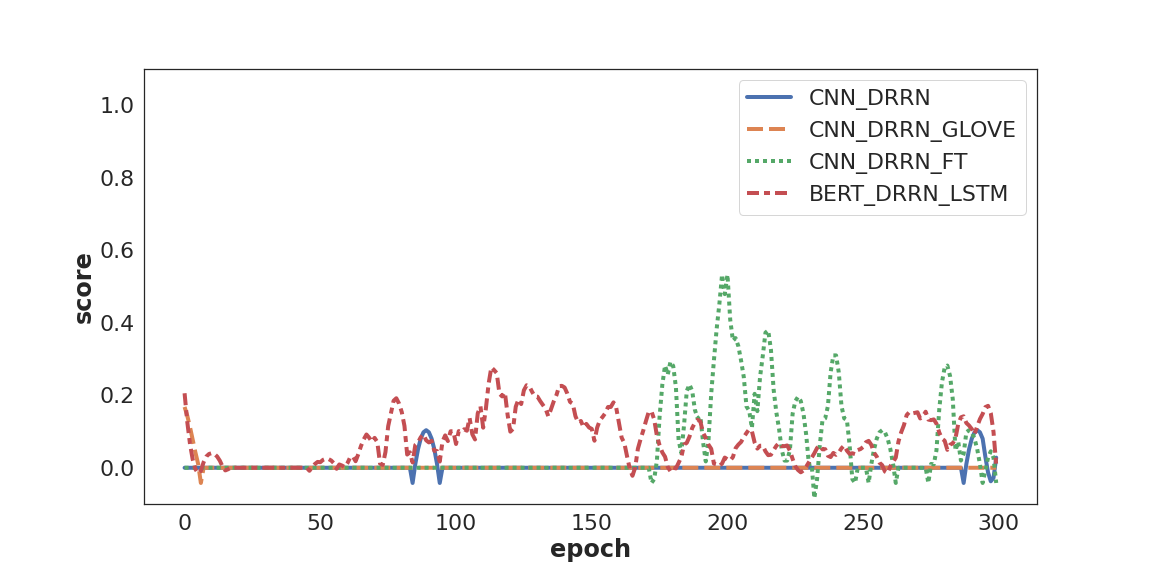}
        \caption{Average evaluation score}
    \end{subfigure}
    \caption{Average scores for CNN encoder with different embedding types}
    \label{figure:7}
\end{figure*}

\begin{figure*}[h]
    \centering
    \begin{subfigure}[b]{0.48\textwidth}
        \centering
        \includegraphics[width=\linewidth]{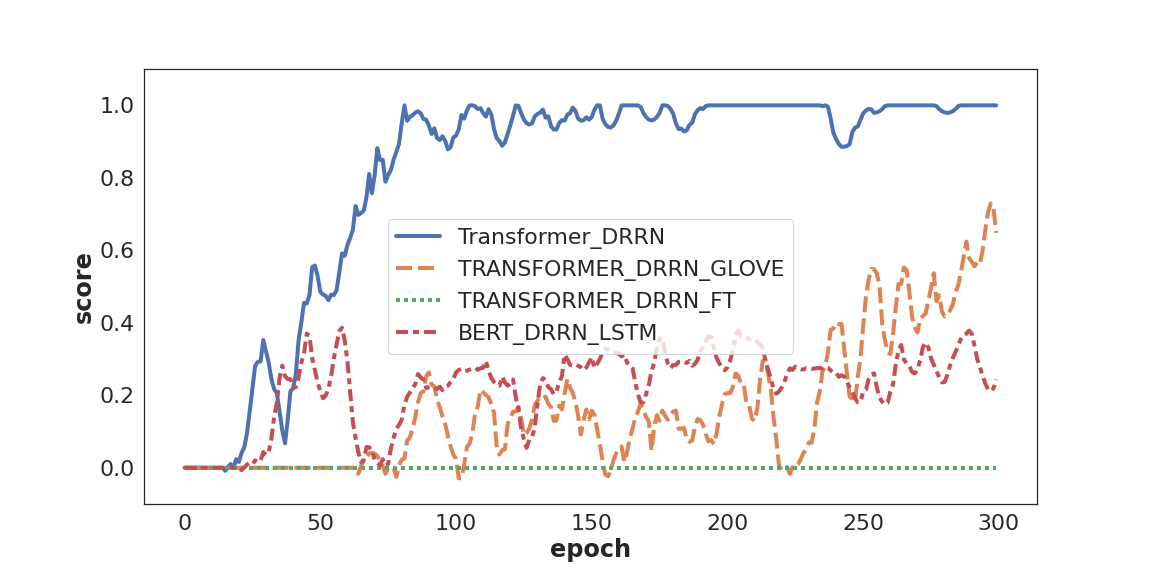}
        \caption{Average training score}
    \end{subfigure}
    \begin{subfigure}[b]{0.48\textwidth}
        \centering
        \includegraphics[width=\linewidth]{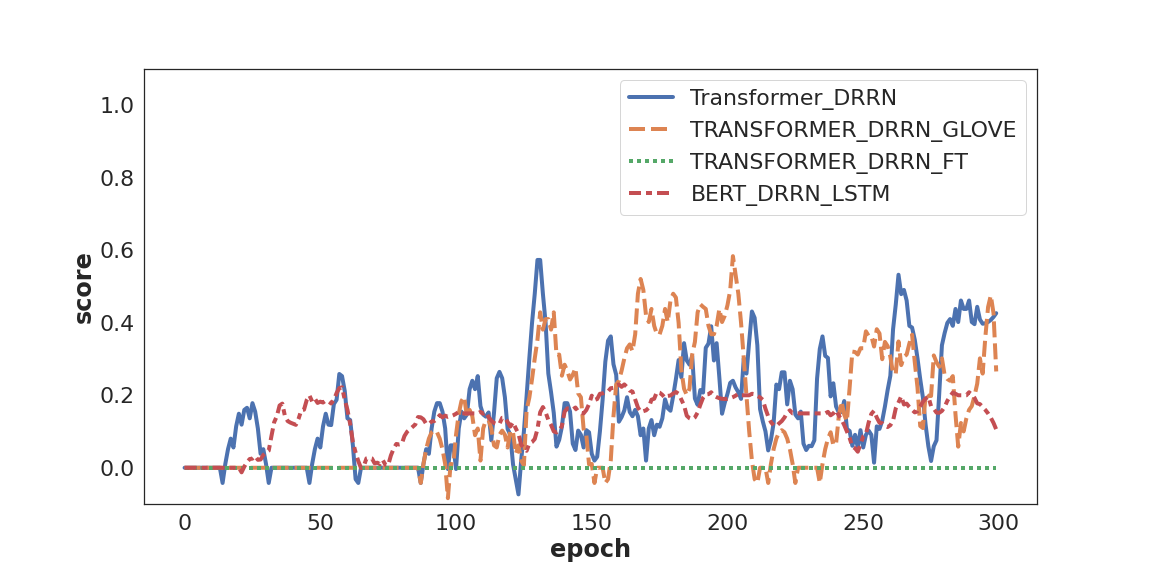}
        \caption{Average evaluation score}
    \end{subfigure}
    \caption{Average scores for Transformer encoder with different embedding types}
    \label{figure:8}
\end{figure*}

Word embeddings contain external information; using word embedding means the model is initialized with external knowledge, and it's important to see if this external knowledge helps model to perform better. To see which type of work embedding is more suitable for TBG tasks, we benchmark three popular pre-trained word embeddings and an embedding layer that is trained in the tasks. For pre-trained embeddings, we include GLOVE, Fasttext, BERT, and see their effect on four encoder types, including GRU, LSTM, CNN and Transformer.

In terms of fitting a large training set of games, from fig.\ref{figure:5}-\ref{figure:8}(a), we can see that the non-pretrained task-specific embedding layer performs much better compared to applying any pretrained word embedding. For all encoder types, the average training score can approach close to 1.0, which means the agent can learn to solve all games in the training set. As text-based game language is generally less complex than traditional NLP tasks such as text classification or translation, embedding trained from a rich corpus does not help the models to learn faster from the simple language environment. Moreover, the non-pretrained embedding layers learn from the rewards the agent receives, which will be more directly related to the environment and crafted reward functions. 

In the training scenario, Glove outperforms the other types among pre-trained embeddings. For all encoder types, Glove outperforms FastText and BERT embedding. BERT embedding is pretrained from corpus larger than Glove and FastText, yet it performs the worst when used with GRU, LSTM and CNN encoder. This finding suggests that language processing methods that perform well in traditional NLP tasks could affect RL agent performance, even used in a language environment. TBG task-specific methods need to be introduced to improve agent performance rather than simply applying traditional NLP methods.

Different embedding also affects the TBG agent model's generalization ability. fig.\ref{figure:5}-\ref{figure:8}(b) presents agents performance on unseen games with different types of word embeddings. By observing fig \ref{figure:5}-\ref{figure:7}(b), for GRU/LSTM/CNN model, FastText and BERT embedding outperforms other embedding types. non-pretrained and Glove embedding perform best in training, but on unseen games, these two types perform poorly, which indicates that they cause the agents to overfit the training set. The different embedding types does not influence the transformer encoder's generalization ability a lot. In fig.\ref{figure:8}(b), all embedding types has similar effect on evaluation score.


\section{Discussion and Future Work} \label{sec slot}

In this work we have summarized methods in Text-based games field, and evaluated selected method on standardized agent and environment and reported our findings. In this section we discuss on some topics which we find important to Text-based game agent design, and offer our perspective on challenges and future directions of Text-based game agent research field.

\subsection{Performance versus efficiency}

In our experiment, we observed that models with more parameters, like BERT, which perform well in traditional NLP tasks such as translation and text classification, doesn't necessarily perform well in TBG tasks. The large models have stronger language encoding capability, but we found that as training the full model is inefficient and requires more complex training techniques, and knowledge acquired from other corpus doesn't necessarily help the model perform better in TBG task, a middle-sized model like single layer transformer encoder or single layer text CNN is more suitable for TBG agent. In order to use large models like BERT, new designs and training techniques need to be introduced in the field to empower the large models.

\subsection{Generalization}

Based on our analysis[6.2.1] and review of related works\cite{yuan2018counting,adhikari2020learning}, model's generalization ability remains a major challenge in this field. The models that can easily learn to solve seen games doesn't guarantee they perform well in unseen game at all. For TBG agent research field, although platforms like Textworld support game set generation, the richness and volume of games are still not comparable to traditional NLP task datasets.For example, a model trained on Wikipedia corpus can have 4.1 billion words with great variety in the training set, but Textworld games training set with 100 games may only have 50 thousands words with very limited vocabulary variations. Because of limited vocabulary and sentence structures in the training set, better structured state representation that generalizes to unseen conditions need to be introduced, to improve generalization ability of the models after trained on limited games.

\subsection{Towards smarter Text-based game agents}

Except for two issues addressed above, TBG agents are mainly trained to evaluate language formed state-action pairs. Among a review of related works[ref ontology] and the analysis of current methods, the agents still lack of learning a general strategy to solve a class of games, especially identifying key game thresholds, like finish gathering a list of different materiel. The agents still lack of the ability to learn and extract useful information from state and action history. The strategy-wise reward functions introduced in the field are limited. Lots of agent designs still contain hand-crafted rules and policies. Hand-crafted rules are also heavily used in game state representation(e.g. graph) generation process. These obstacles remain to be research problems in this field, in order to design a smart agent that solves the games as efficiently, as resourceful as human.

\section{Conclusion}

In this paper, we choose some RNN, CNN, and Transformer based models to represent popularly used models in TBG and NLP fields. The selected models also contain different levels of total trainable parameters to represent different levels of efficiency. We train the models using the same hyper-parameters, with the same training games set and evaluate model game performance both on seen and unseen games. The results show that CNN and Transformer perform best, and Transformer has a much lower parameter count. The models with more parameters like BERT do not deliver stronger performance. The models do not solve all the seen games, even for easy games. For hard games like cook, the best performing models can only solve about 50\% of the seen games. For unseen games, the models perform poorly. To reduce training fluctuation, better training strategies need to be introduced to TBG games. For RL algorithms like Actor-Critic, how to better apply it to TBG agent training to improve training performance remains to be found. Current applications of Actor-Critic in TBG agent training don't bring significant benefits. For harder games, better reward functions should be designed to help the model value key thresholds to finish the games.

We standardise our environment and training procedure and remove all hand-crafted rules, to present how the modules and word embedding types contribute toward agent performance. The language processing modules and RL components are also standardised. We pick games with different difficulties and form the same training sets and generalisation evaluation sets. We see that, without adding hand-crafted rules and other policy-related pretrained components, some modules like the Transformer encoder and non-pretrained embedding perform better. The modules that contain external information, like word embeddings, do not really help the agents' performance as expected. As applying pre-trained modules that work in traditional NLP tasks does not necessarily enhance the agent performance, external knowledge and pre-trained modules need to be tuned better to fit TBG agent training.

We have chosen two evaluation types-multiple game tests and generalisation tests to see if models can perform consistently in different evaluation types. These two types are more meaningful for TBG agent applications. We train the models using a set of games and evaluate them on seen and unseen game sets. The result shows that the model's performance in multiple and generalisation tests has a very weak correlation. Those models that perform better on seen games do not generalise well. The state representation is not well structured to generalise to unseen states. Thus the state value approximation only works for seen games. A better-structured state representation design with a smaller feature space needs to be introduced to represent more states and improve model generalisation performance. It's also important to try to train the agent to work on different game types, for example, to train the agent to learn from Coin-collector and Treasure hunter, and examine its ability to solve the Cooking game. The learnt knowledge should be transferable between a larger variety of text-based games.

Through our tests analysis, we present the strengths and weaknesses of selected deep learning components used in TBG research works. Recent developments in the TBG agent field mainly focus on graph representation and external modules to enhance the models' performance. Yet the model generalisation to unseen games and model capability remain major challenges in this field. As a reference for future research works, new methods could be introduced to cope with the strengths and weaknesses of the deep learning modules analysed in our work.

\acks{The authors wish to thank Hans-Martin Adorf, Don Rosenthal, 
Richard Franier, Peter Cheeseman and Monte Zweben for their assistance
and advice.  We also thank Ron Musick and our anonymous reviewers for
their comments.  The Space Telescope Science Institute is operated by
the Association of Universities for Research in Astronomy for NASA.
}

\appendix
\section*{Appendix A. Probability Distributions for N-Queens}

[section ommitted]

\vskip 0.2in
\bibliographystyle{theapa.bst}
\bibliography{tbg_bib}

\begin{thebibliography}{}

\bibitem[\protect\BCAY{Adhikari, Yuan, C{\^o}t{\'e}, Zelinka, Rondeau, Laroche,
  Poupart, Tang, Trischler,\ \BBA\ Hamilton}{Adhikari
  et~al.}{2020}]{adhikari2020learning}
Adhikari, A., Yuan, X., C{\^o}t{\'e}, M.-A., Zelinka, M., Rondeau, M.-A.,
  Laroche, R., Poupart, P., Tang, J., Trischler, A., \BBA\ Hamilton, W.
  \BBOP2020\BBCP.
\newblock \BBOQ Learning dynamic belief graphs to generalize on text-based
  games\BBCQ\
\newblock {\Bem Advances in Neural Information Processing Systems}, {\Bem 33}.

\bibitem[\protect\BCAY{Adolphs\ \BBA\ Hofmann}{Adolphs\ \BBA\
  Hofmann}{2020}]{adolphs2020ledeepchef}
Adolphs, L.\BBACOMMA\  \BBA\ Hofmann, T. \BBOP2020\BBCP.
\newblock \BBOQ Ledeepchef deep reinforcement learning agent for families of
  text-based games\BBCQ\
\newblock In {\Bem Proceedings of the AAAI Conference on Artificial
  Intelligence}, \lowercase{\BVOL}~34, \BPGS\ 7342--7349.

\bibitem[\protect\BCAY{Ammanabrolu\ \BBA\ Hausknecht}{Ammanabrolu\ \BBA\
  Hausknecht}{2019}]{ammanabrolu2019graph}
Ammanabrolu, P.\BBACOMMA\  \BBA\ Hausknecht, M. \BBOP2019\BBCP.
\newblock \BBOQ Graph constrained reinforcement learning for natural language
  action spaces\BBCQ\
\newblock In {\Bem International Conference on Learning Representations}.

\bibitem[\protect\BCAY{Ammanabrolu\ \BBA\ Riedl}{Ammanabrolu\ \BBA\
  Riedl}{2019}]{ammanabrolu2019playing}
Ammanabrolu, P.\BBACOMMA\  \BBA\ Riedl, M. \BBOP2019\BBCP.
\newblock \BBOQ Playing text-adventure games with graph-based deep
  reinforcement learning\BBCQ\
\newblock In {\Bem NAACL-HLT (1)}.

\bibitem[\protect\BCAY{Ammanabrolu\ \BBA\ Riedl}{Ammanabrolu\ \BBA\
  Riedl}{2021}]{ammanabrolu2021learning}
Ammanabrolu, P.\BBACOMMA\  \BBA\ Riedl, M. \BBOP2021\BBCP.
\newblock \BBOQ Learning knowledge graph-based world models of textual
  environments\BBCQ\
\newblock {\Bem Advances in Neural Information Processing Systems}, {\Bem 34},
  3720--3731.

\bibitem[\protect\BCAY{Ammanabrolu, Tien, Hausknecht,\ \BBA\ Riedl}{Ammanabrolu
  et~al.}{2020}]{ammanabrolu2020avoid}
Ammanabrolu, P., Tien, E., Hausknecht, M., \BBA\ Riedl, M.~O. \BBOP2020\BBCP.
\newblock \BBOQ How to avoid being eaten by a grue: Structured exploration
  strategies for textual worlds\BBCQ\
\newblock {\Bem arXiv preprint arXiv:2006.07409}, {\Bem 1\/}(1).

\bibitem[\protect\BCAY{Chen, Fisch, Weston,\ \BBA\ Bordes}{Chen
  et~al.}{2017}]{chen2017reading}
Chen, D., Fisch, A., Weston, J., \BBA\ Bordes, A. \BBOP2017\BBCP.
\newblock \BBOQ Reading wikipedia to answer open-domain questions\BBCQ.

\bibitem[\protect\BCAY{C{\^o}t{\'e}, K{\'a}d{\'a}r, Yuan, Kybartas, Barnes,
  Fine, Moore, Hausknecht, El~Asri, Adada, et~al.}{C{\^o}t{\'e}
  et~al.}{2018}]{cote2018textworld}
C{\^o}t{\'e}, M.-A., K{\'a}d{\'a}r, {\'A}., Yuan, X., Kybartas, B., Barnes, T.,
  Fine, E., Moore, J., Hausknecht, M., El~Asri, L., Adada, M., et~al.
  \BBOP2018\BBCP.
\newblock \BBOQ Textworld: A learning environment for text-based games\BBCQ\
\newblock In {\Bem Workshop on Computer Games}, \BPGS\ 41--75. Springer.

\bibitem[\protect\BCAY{Hausknecht, Ammanabrolu, C{\^o}t{\'e},\ \BBA\
  Yuan}{Hausknecht et~al.}{2020}]{hausknecht2020interactive}
Hausknecht, M., Ammanabrolu, P., C{\^o}t{\'e}, M.-A., \BBA\ Yuan, X.
  \BBOP2020\BBCP.
\newblock \BBOQ Interactive fiction games: A colossal adventure\BBCQ\
\newblock In {\Bem Proceedings of the AAAI Conference on Artificial
  Intelligence}, \lowercase{\BVOL}~34, \BPGS\ 7903--7910.

\bibitem[\protect\BCAY{Hausknecht, Ammanabrolu, Marc-Alexandre,\ \BBA\
  Xingdi}{Hausknecht et~al.}{2019}]{hausknecht19}
Hausknecht, M., Ammanabrolu, P., Marc-Alexandre, C., \BBA\ Xingdi, Y.
  \BBOP2019\BBCP.
\newblock \BBOQ Interactive fiction games: A colossal adventure\BBCQ\
\newblock {\Bem CoRR}, {\Bem abs/1909.05398}.

\bibitem[\protect\BCAY{He, Chen, He, Gao, Li, Deng,\ \BBA\ Ostendorf}{He
  et~al.}{2015}]{he2015deep}
He, J., Chen, J., He, X., Gao, J., Li, L., Deng, L., \BBA\ Ostendorf, M.
  \BBOP2015\BBCP.
\newblock \BBOQ Deep reinforcement learning with an unbounded action
  space\BBCQ\
\newblock {\Bem arXiv preprint arXiv:1511.04636}, {\Bem 5}.

\bibitem[\protect\BCAY{Honnibal, Montani, Van~Landeghem,\ \BBA\ Boyd}{Honnibal
  et~al.}{2020}]{Honnibal_spaCy_Industrial-strength_Natural_2020}
Honnibal, M., Montani, I., Van~Landeghem, S., \BBA\ Boyd, A. \BBOP2020\BBCP.
\newblock \BBOQ {spaCy: Industrial-strength Natural Language Processing in
  Python}\BBCQ.
\newblock {\Bem 1\/}(1).

\bibitem[\protect\BCAY{Jack~Urbanek}{Jack~Urbanek}{2019}]{urbanek2019light}
Jack~Urbanek, A. F. e.~a. \BBOP2019\BBCP.
\newblock \BBOQ Learning to speak and act in a fantasy text adventure
  game\BBCQ.

\bibitem[\protect\BCAY{Jain, Fedus, Larochelle, Precup,\ \BBA\ Bellemare}{Jain
  et~al.}{2020}]{jain2020algorithmic}
Jain, V., Fedus, W., Larochelle, H., Precup, D., \BBA\ Bellemare, M.~G.
  \BBOP2020\BBCP.
\newblock \BBOQ Algorithmic improvements for deep reinforcement learning
  applied to interactive fiction.\BBCQ\
\newblock In {\Bem AAAI}, \BPGS\ 4328--4336.

\bibitem[\protect\BCAY{Jang, Seo, Lee,\ \BBA\ Kim}{Jang
  et~al.}{2020}]{jang2020monte}
Jang, Y., Seo, S., Lee, J., \BBA\ Kim, K.-E. \BBOP2020\BBCP.
\newblock \BBOQ Monte-carlo planning and learning with language action value
  estimates\BBCQ\
\newblock In {\Bem International Conference on Learning Representations}.

\bibitem[\protect\BCAY{Kenton\ \BBA\ Toutanova}{Kenton\ \BBA\
  Toutanova}{2019}]{kenton2019bert}
Kenton, J. D. M.-W.~C.\BBACOMMA\  \BBA\ Toutanova, L.~K. \BBOP2019\BBCP.
\newblock \BBOQ Bert: Pre-training of deep bidirectional transformers for
  language understanding\BBCQ\
\newblock In {\Bem Proceedings of NAACL-HLT}, \BPGS\ 4171--4186.

\bibitem[\protect\BCAY{Kim}{Kim}{2014}]{kim-2014-convolutional}
Kim, Y. \BBOP2014\BBCP.
\newblock \BBOQ Convolutional neural networks for sentence classification\BBCQ\
\newblock In {\Bem Proceedings of the 2014 Conference on Empirical Methods in
  Natural Language Processing ({EMNLP})}, \BPGS\ 1746--1751, Doha, Qatar.
  Association for Computational Linguistics.

\bibitem[\protect\BCAY{Lan, Chen, Goodman, Gimpel, Sharma,\ \BBA\ Soricut}{Lan
  et~al.}{2019}]{lan2019albert}
Lan, Z., Chen, M., Goodman, S., Gimpel, K., Sharma, P., \BBA\ Soricut, R.
  \BBOP2019\BBCP.
\newblock \BBOQ Albert: A lite bert for self-supervised learning of language
  representations\BBCQ\
\newblock In {\Bem International Conference on Learning Representations}.

\bibitem[\protect\BCAY{Mikolov, Chen, Corrado,\ \BBA\ Dean}{Mikolov
  et~al.}{2013}]{mikolov2013efficient}
Mikolov, T., Chen, K., Corrado, G., \BBA\ Dean, J. \BBOP2013\BBCP.
\newblock \BBOQ Efficient estimation of word representations in vector
  space\BBCQ\
\newblock {\Bem arXiv preprint arXiv:1301.3781}, {\Bem 1\/}(1).

\bibitem[\protect\BCAY{Mikolov, Grave, Bojanowski, Puhrsch,\ \BBA\
  Joulin}{Mikolov et~al.}{2018}]{mikolov2018advances}
Mikolov, T., Grave, {\'E}., Bojanowski, P., Puhrsch, C., \BBA\ Joulin, A.
  \BBOP2018\BBCP.
\newblock \BBOQ Advances in pre-training distributed word representations\BBCQ\
\newblock In {\Bem Proceedings of the Eleventh International Conference on
  Language Resources and Evaluation (LREC 2018)}.

\bibitem[\protect\BCAY{Murugesan, Atzeni, Kapanipathi, Talamadupula, Sachan,\
  \BBA\ Campbell}{Murugesan et~al.}{2021}]{murugesan2021efficient}
Murugesan, K., Atzeni, M., Kapanipathi, P., Talamadupula, K., Sachan, M., \BBA\
  Campbell, M. \BBOP2021\BBCP.
\newblock \BBOQ Efficient text-based reinforcement learning by jointly
  leveraging state and commonsense graph representations\BBCQ\
\newblock In {\Bem Acl-Ijcnlp 2021: The 59Th Annual Meeting Of The Association
  For Computational Linguistics And The 11Th International Joint Conference On
  Natural Language Processing, Vol 2}, \BPGS\ 719--725. ASSOC COMPUTATIONAL
  LINGUISTICS-ACL.

\bibitem[\protect\BCAY{Murugesan, Atzeni, Shukla, Sachan, Kapanipathi,\ \BBA\
  Talamadupula}{Murugesan et~al.}{2020}]{murugesan2020enhancing}
Murugesan, K., Atzeni, M., Shukla, P., Sachan, M., Kapanipathi, P., \BBA\
  Talamadupula, K. \BBOP2020\BBCP.
\newblock \BBOQ Enhancing text-based reinforcement learning agents with
  commonsense knowledge\BBCQ\
\newblock {\Bem arXiv preprint arXiv:2005.00811}, {\Bem 1\/}(1).

\bibitem[\protect\BCAY{Nahian, Frazier, Harrison,\ \BBA\ Riedl}{Nahian
  et~al.}{2021}]{nahian2021training}
Nahian, M. S.~A., Frazier, S., Harrison, B., \BBA\ Riedl, M. \BBOP2021\BBCP.
\newblock \BBOQ Training value-aligned reinforcement learning agents using a
  normative prior\BBCQ\
\newblock {\Bem arXiv preprint arXiv:2104.09469}, {\Bem 1\/}(1).

\bibitem[\protect\BCAY{Narasimhan, Kulkarni,\ \BBA\ Barzilay}{Narasimhan
  et~al.}{2015}]{narasimhan2015language}
Narasimhan, K., Kulkarni, T.~D., \BBA\ Barzilay, R. \BBOP2015\BBCP.
\newblock \BBOQ Language understanding for textbased games using deep
  reinforcement learning\BBCQ\
\newblock In {\Bem In Proceedings of the Conference on Empirical Methods in
  Natural Language Processing}. Citeseer.

\bibitem[\protect\BCAY{Pennington, Socher,\ \BBA\ Manning}{Pennington
  et~al.}{2014}]{pennington2014glove}
Pennington, J., Socher, R., \BBA\ Manning, C.~D. \BBOP2014\BBCP.
\newblock \BBOQ Glove: Global vectors for word representation\BBCQ\
\newblock In {\Bem Proceedings of the 2014 conference on empirical methods in
  natural language processing (EMNLP)}, \BPGS\ 1532--1543.

\bibitem[\protect\BCAY{Radford, Wu, Child, Luan, Amodei, Sutskever,
  et~al.}{Radford et~al.}{2019}]{radford2019language}
Radford, A., Wu, J., Child, R., Luan, D., Amodei, D., Sutskever, I., et~al.
  \BBOP2019\BBCP.
\newblock \BBOQ Language models are unsupervised multitask learners\BBCQ\
\newblock {\Bem OpenAI blog}, {\Bem 1\/}(8), 9.

\bibitem[\protect\BCAY{Savitzky\ \BBA\ Golay}{Savitzky\ \BBA\
  Golay}{1964}]{savitzky1964smoothing}
Savitzky, A.\BBACOMMA\  \BBA\ Golay, M.~J. \BBOP1964\BBCP.
\newblock \BBOQ Smoothing and differentiation of data by simplified least
  squares procedures.\BBCQ\
\newblock {\Bem Analytical chemistry}, {\Bem 36\/}(8), 1627--1639.

\bibitem[\protect\BCAY{Simonyan\ \BBA\ Zisserman}{Simonyan\ \BBA\
  Zisserman}{}]{simonyan2015deep}
Simonyan, K.\BBACOMMA\  \BBA\ Zisserman, A.
\newblock \BBOQ Very deep convolutional networks for large-scale image
  recognition\BBCQ\
\newblock In {\Bem ICLR 2015}.

\bibitem[\protect\BCAY{Szegedy, Liu, Jia, Sermanet, Reed, Anguelov, Erhan,
  Vanhoucke,\ \BBA\ Rabinovich}{Szegedy et~al.}{2015}]{szegedy2015going}
Szegedy, C., Liu, W., Jia, Y., Sermanet, P., Reed, S., Anguelov, D., Erhan, D.,
  Vanhoucke, V., \BBA\ Rabinovich, A. \BBOP2015\BBCP.
\newblock \BBOQ Going deeper with convolutions\BBCQ\
\newblock In {\Bem Proceedings of the IEEE conference on computer vision and
  pattern recognition}, \BPGS\ 1--9.

\bibitem[\protect\BCAY{Vaswani, Shazeer, Parmar, Uszkoreit, Jones, Gomez,
  Kaiser,\ \BBA\ Polosukhin}{Vaswani et~al.}{2017}]{vaswani2017attention}
Vaswani, A., Shazeer, N., Parmar, N., Uszkoreit, J., Jones, L., Gomez, A.~N.,
  Kaiser, {\L}., \BBA\ Polosukhin, I. \BBOP2017\BBCP.
\newblock \BBOQ Attention is all you need\BBCQ\
\newblock In {\Bem Advances in neural information processing systems}, \BPGS\
  5998--6008.

\bibitem[\protect\BCAY{Xu, Fang, Chen, Du, Zhou,\ \BBA\ Zhang}{Xu
  et~al.}{2020}]{xu2020deep2}
Xu, Y., Fang, M., Chen, L., Du, Y., Zhou, J.~T., \BBA\ Zhang, C.
  \BBOP2020\BBCP.
\newblock \BBOQ Deep reinforcement learning with stacked hierarchical attention
  for text-based games\BBCQ\
\newblock {\Bem Advances in Neural Information Processing Systems}, {\Bem 33}.

\bibitem[\protect\BCAY{Yao, Narasimhan,\ \BBA\ Hausknecht}{Yao
  et~al.}{2021}]{yao2021reading}
Yao, S., Narasimhan, K., \BBA\ Hausknecht, M. \BBOP2021\BBCP.
\newblock \BBOQ Reading and acting while blindfolded: The need for semantics in
  text game agents\BBCQ\
\newblock In {\Bem Proceedings of the 2021 Conference of the North American
  Chapter of the Association for Computational Linguistics: Human Language
  Technologies}, \BPGS\ 3097--3102.

\bibitem[\protect\BCAY{Yao, Rao, Hausknecht,\ \BBA\ Narasimhan}{Yao
  et~al.}{2020}]{yao2020keep}
Yao, S., Rao, R., Hausknecht, M., \BBA\ Narasimhan, K. \BBOP2020\BBCP.
\newblock \BBOQ Keep calm and explore: Language models for action generation in
  text-based games\BBCQ\
\newblock In {\Bem Proceedings of the 2020 Conference on Empirical Methods in
  Natural Language Processing (EMNLP)}, \BPGS\ 8736--8754.

\bibitem[\protect\BCAY{Yin\ \BBA\ May}{Yin\ \BBA\
  May}{2019a}]{yin2019comprehensible}
Yin, X.\BBACOMMA\  \BBA\ May, J. \BBOP2019a\BBCP.
\newblock \BBOQ Comprehensible context-driven text game playing\BBCQ\
\newblock In {\Bem 2019 IEEE Conference on Games (CoG)}, \BPGS\ 1--8. IEEE.

\bibitem[\protect\BCAY{Yin\ \BBA\ May}{Yin\ \BBA\ May}{2019b}]{yin2019learn}
Yin, X.\BBACOMMA\  \BBA\ May, J. \BBOP2019b\BBCP.
\newblock \BBOQ Learn how to cook a new recipe in a new house: Using map
  familiarization, curriculum learning, and bandit feedback to learn families
  of text-based adventure games\BBCQ\
\newblock {\Bem arXiv preprint arXiv:1908.04777}, {\Bem 1\/}(1).

\bibitem[\protect\BCAY{Yin, Weischedel,\ \BBA\ May}{Yin
  et~al.}{2020}]{yin2020learning}
Yin, X., Weischedel, R., \BBA\ May, J. \BBOP2020\BBCP.
\newblock \BBOQ Learning to generalize for sequential decision making\BBCQ\
\newblock In {\Bem Findings of the Association for Computational Linguistics:
  EMNLP 2020}, \BPGS\ 3046--3063.

\bibitem[\protect\BCAY{Yuan, C{\^o}t{\'e}, Fu, Lin, Pal, Bengio,\ \BBA\
  Trischler}{Yuan et~al.}{2019}]{yuan2019interactive}
Yuan, X., C{\^o}t{\'e}, M.-A., Fu, J., Lin, Z., Pal, C., Bengio, Y., \BBA\
  Trischler, A. \BBOP2019\BBCP.
\newblock \BBOQ Interactive language learning by question answering\BBCQ\
\newblock In {\Bem Proceedings of the 2019 Conference on Empirical Methods in
  Natural Language Processing and the 9th International Joint Conference on
  Natural Language Processing (EMNLP-IJCNLP)}, \BPGS\ 2796--2813.

\bibitem[\protect\BCAY{Yuan, C{\^o}t{\'e}, Sordoni, Laroche, Combes,
  Hausknecht,\ \BBA\ Trischler}{Yuan et~al.}{2018}]{yuan2018counting}
Yuan, X., C{\^o}t{\'e}, M.-A., Sordoni, A., Laroche, R., Combes, R. T.~d.,
  Hausknecht, M., \BBA\ Trischler, A. \BBOP2018\BBCP.
\newblock \BBOQ Counting to explore and generalize in text-based games\BBCQ\
\newblock In {\Bem ICML 2018}.

\bibitem[\protect\BCAY{Zahavy, Haroush, Merlis, Mankowitz,\ \BBA\
  Mannor}{Zahavy et~al.}{2018}]{zahavy2018learn}
Zahavy, T., Haroush, M., Merlis, N., Mankowitz, D.~J., \BBA\ Mannor, S.
  \BBOP2018\BBCP.
\newblock \BBOQ Learn what not to learn: Action elimination with deep
  reinforcement learning\BBCQ\
\newblock In {\Bem Advances in Neural Information Processing Systems}, \BPGS\
  3562--3573.

\end{thebibliography}

\end{document}